\documentclass[10pt,twocolumn,letterpaper]{article}

\usepackage{iccv}
\usepackage{times}
\usepackage{epsfig}
\usepackage{graphicx}
\usepackage{amsmath}
\usepackage{amssymb}
\usepackage{booktabs}
\usepackage{bbding}
\usepackage{multirow}
\usepackage{subfigure}
\usepackage{color,xcolor}
\usepackage{pifont}
\usepackage{subfigure}
\usepackage{colortbl}

\usepackage[accsupp]{axessibility}  

\usepackage[pagebackref=true,breaklinks=true,letterpaper=true,colorlinks,bookmarks=false]{hyperref}

\iccvfinalcopy 

\def\iccvPaperID{****} 

\ificcvfinal\fi

\begin{document}

\title{TransPose: Keypoint Localization via Transformer}

\author{Sen Yang \quad Zhibin Quan \quad Mu Nie \quad Wankou Yang\thanks{Corresponding author.}\\[0.5mm]
School of Automation, Southeast University, Nanjing 210096, China\\
	{\tt\small \{yangsenius, 101101872, niemu, wkyang\}@seu.edu.cn}
}

\maketitle
\maketitle
\ificcvfinal\thispagestyle{empty}\fi

	\begin{abstract}
	While CNN-based models have made remarkable progress on human pose estimation, what spatial dependencies they capture to localize keypoints remains unclear. In this work, we propose a model called \textbf{TransPose}, which introduces Transformer for human pose estimation. The attention layers built in Transformer enable our model to capture long-range relationships efficiently and also can reveal what dependencies the predicted keypoints rely on. To predict keypoint heatmaps, the last attention layer acts as an aggregator, which collects contributions from image clues and forms maximum positions of keypoints. Such a heatmap-based localization approach via Transformer conforms to the principle of Activation Maximization~\cite{erhan2009visualizing}. And the revealed dependencies are image-specific and fine-grained, which also can provide evidence of how the model handles special cases, e.g., occlusion. The experiments show that TransPose achieves 75.8 AP and 75.0 AP on COCO validation and test-dev sets, while being more lightweight and faster than mainstream CNN architectures. The TransPose model also transfers very well on MPII benchmark, achieving superior performance on the test set when fine-tuned with small training costs. Code and pre-trained models are publicly available\footnote{ \url{https://github.com/yangsenius/TransPose}}.

\end{abstract}\vspace*{-0.2in}

\section{Introduction}

\begin{figure}[t]
	\begin{center}
		\includegraphics[width=0.8\linewidth]{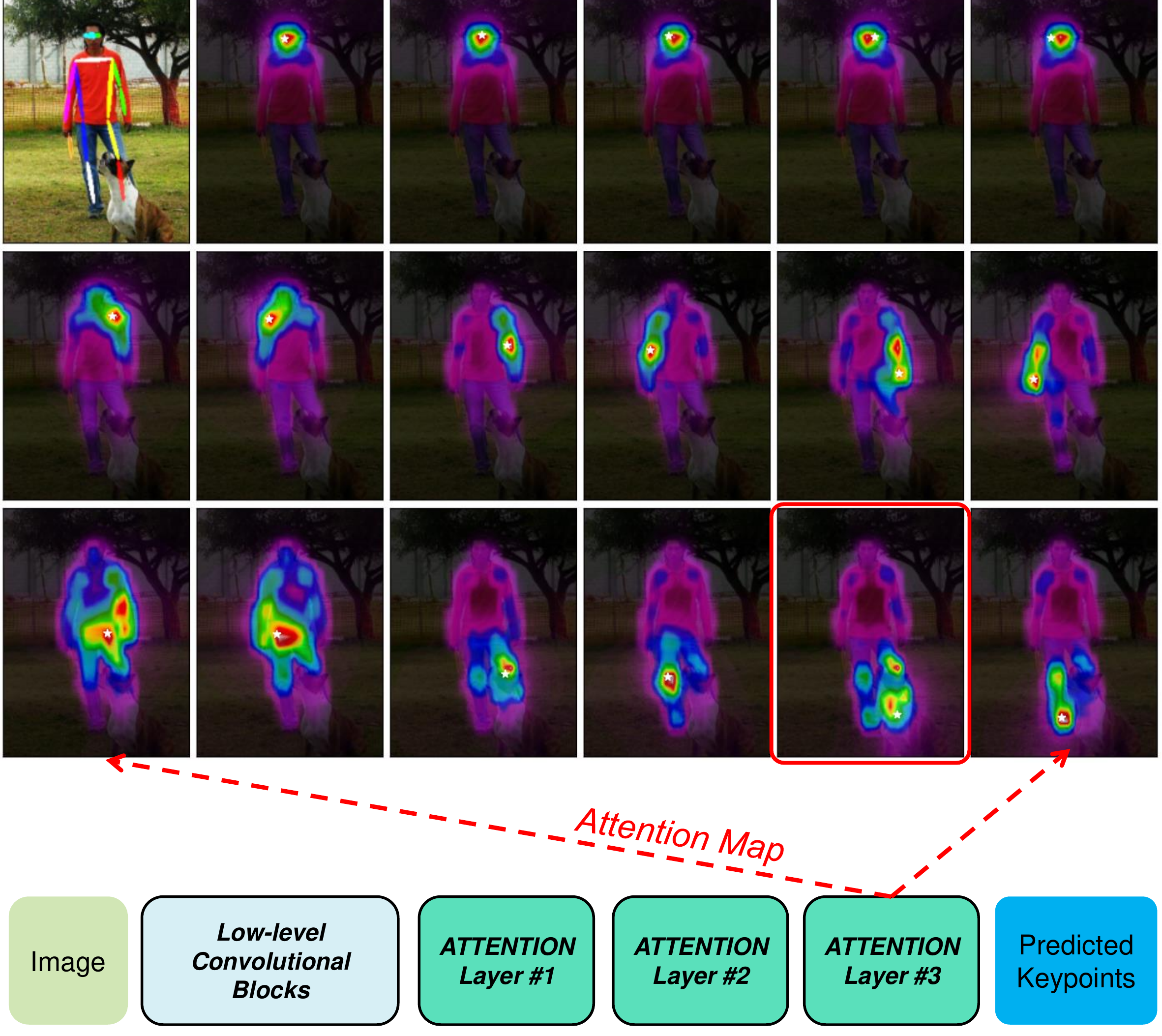}
	\end{center}
	\caption{A schematic diagram of TransPose. \textbf{Below:} The inference pipeline. \textbf{Above:}  Dependency areas for each predicted keypoint location. In this example, the person's left-ankle is occluded by a dog. \textbf{\emph{Which exact image clues the model uses to infer the occluded joint?}} The attention map (red box) gives \emph{fine-grained evidence beyond intuition}: such a pose estimator highly relies on the image clues around the left ankle, left upper leg, and joints on the right leg to estimate the location of occluded left ankle.}\vspace*{-0.1in}
	\label{beginning fig}
\end{figure}
Deep convolutional neural networks have achieved impressive performances in the field of human pose estimation. 
DeepPose~\cite{toshev2014deeppose} is the early classic method, directly regressing the numerical coordinate locations of keypoints. 
Afterwards, fully convolutional networks like ~\cite{wei2016convolutional,long2015fully, newell2016stacked,yang2017learning,chen2018cascaded,papandreou2018PersonLabPP,xiao2018simple,sun2019hrnet} have become the mainstream by predicting keypoints heatmaps, which \emph{implicitly} learn spatial dependencies between body parts. 
Yet, most prior works take deep CNN as a powerful black box predictor and focus on improving the network structure, what exactly happens inside the models or how they capture the spatial relationships between body parts remains unclear. 
However, from the scientific and practical standpoints, the interpretability of the model can aid practitioners the ability to understand how the model associates structural variables to reach the final predictions and how a pose estimator handles various input images. 
It also can help model developers for debugging, decision-making, and further improving the design.

For existing pose estimators, some issues make it challenging to figure out their decision processes. 
\emph{(1) Deepness}. The CNN-based models, such as~\cite{wei2016convolutional,newell2016stacked,xiao2018simple,sun2019hrnet}, are usually very deep non-linear models that hinder the interpretation of the function of each layer. 
\emph{(2) Implicit relationships}. The global spatial relationships between body parts are implicitly encoded within the neuron activations and the weights of CNNs. It is not easy to decouple such relationships from large amounts of weights and activations in neural networks. 
And solely visualizing the intermediate features with a large number of channels (e.g. 256, 512 in SimpleBaseline architecture~\cite{xiao2018simple}) provides little meaningful explanations. 
\emph{(3) Limited working memory in inferring various images}. The desired explanations for the model predictions should be image-specific and fine-grained. 
When inferring images, however, the \emph{static} convolution kernels are limited in the ability to represent variables due to the limited working memory~\cite{graves2014neural,graves2016hybrid, hochreiter1997long}. 
So it is difficult for CNNs to capture image-specific dependencies due to their content-independent parameters yet variable input image contents. 
\emph{(4) Lack of tools.} Although there are already many visualization techniques based on gradient or attribution~\cite{erhan2009visualizing, zeiler2014visualizing, simonyan2013deep, selvaraju2017grad, fong2017interpretable, olah2017feature, zhou2016learning, bach2015pixel}, most of them focus on image classification rather than localization. 
They aim to reveal class-specific input patterns or saliency maps rather than to explain the relationships between structure variables (\emph{e.g.}, the locations of keypoints). By far, how to develop explainable pose estimators remains challenging.

\begin{figure}[t]
	\begin{center}
		\includegraphics[width=0.8\linewidth]{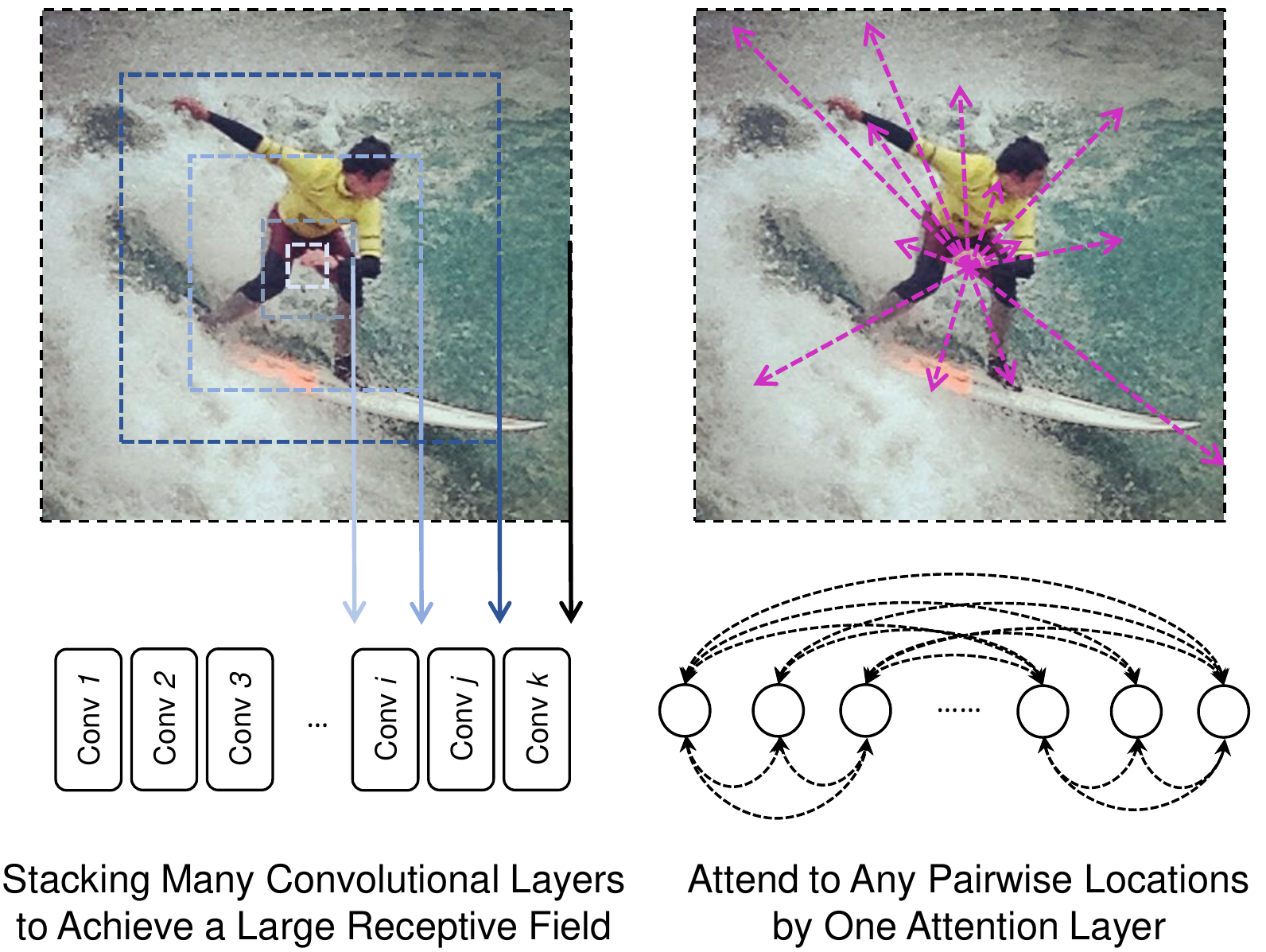}
	\end{center}
	\caption{CNN vs. Attention. \textbf{Left:} The receptive filed enlarges in the deeper convolutional layer. \textbf{Right:} One self-attention layer can capture the pairwise relationship between any pair of locations.}
	\label{conv_vs_attention}\vspace*{-0.1in}
	
\end{figure}

In this work, we aim to build a human pose estimator that can explicitly capture and reveal the image-specific spatial dependencies between keypoints, as shown in Fig.~\ref{beginning fig}. Due to the poor scaling property of convolution~\cite{ramachandran2019stand}, we argue that convolution has advantages in extracting low-level features, but deeply stacking convolutions at high-level to enlarge the receptive field is not efficient to capture global dependencies. And such deepness increases the difficulty in interpreting CNN predictions. Transformer architecture~\cite{vaswani2017attention} has a natural advantage over CNNs in terms of drawing pairwise or higher-order interactions. As shown in Fig.~\ref{conv_vs_attention}, attention layers enable the model to capture interactions between any pairwise locations, and its attention map acts as an immediate memory to store these dependencies. 

Based on these considerations, we propose a novel model called \emph{TransPose}, using convolutions to extract features at low-level and Transformer to capture global dependencies at high-level. In detail, we flatten the feature maps as input to Transformer and recover its output into the 2D-structure heatmaps. In such a design, the last attention layer in Transformer specially acts as an \emph{aggregator}, which collects different contributions from all image locations by attention scores and finally forms the maximum positions in the heatmaps. This type of keypoint localization approach via Transformer establishes a connection with the interpretability of Activation Maximization~\cite{erhan2009visualizing, simonyan2013deep} and extends it to the localization task. The resulting attention scores can indicate what concrete image clues significantly contribute to the predicted locations. With such evidence, we can further analyze the behaviors of the model by examining the influence of different experimental variables. In summary, our contributions are as follow:

\begin{itemize}
	\item We introduce Transformer for human pose estimation to predict heatmap-based keypoints positions, which can efficiently capture the spatial relationships between human body parts. \vspace*{-0.05in}
	
	\item We demonstrate that our keypoint localization approach based on Transformer conforms to the interpretability of Activation Maximization~\cite{erhan2009visualizing, simonyan2013deep}. Qualitative analysis reveals the dependencies beyond intuition, which are image-specific and fine-grained.\vspace*{-0.05in}
	
	\item TransPose models achieve competitive performances against state-of-the-arts CNN-based models via fewer parameters and faster speeds. TransPose achieves 75.8 AP and 75.0 AP on COCO validation set and test-dev set, with 73$\%$ fewer parameters and 1.4$\times$ faster than HRNet-W48. In addition, our model transfers very well on MPII benchmark.  
\end{itemize}

\section{Related Work}

\subsection{Human Pose Estimation}
Deep CNNs have achieved great success in human pose estimation. The inductive biases of vanilla convolution kernel~\cite{lecun1998gradient, krizhevsky2012imagenet} are locality and translation equivariance. It proves to be efficient to extract low-level image feature. For human pose estimation, capturing global dependencies is crucial~\cite{ramakrishna2014pose,tompson2014joint,wei2016convolutional,papandreou2018PersonLabPP}, but the locality nature of convolution makes it impossible to capture long-range interactions. A typical but brute solution is to enlarge the receptive field, \emph{e.g.} by downsampling the resolution, increasing the depth or expanding the kernel size. Further, sophisticated strategies are proposed such as multi-scale fusion~\cite{newell2016stacked, pfister2015flowing, yang2017learning, chen2018cascaded, sun2019hrnet, chu2017multi, cheng2020higher}, stacking~\cite{wei2016convolutional,xiao2018simple, newell2016stacked}, or high-resolution representation~\cite{sun2019hrnet}; meanwhile, many successful architectures have emerged such as CPM~\cite{wei2016convolutional}, Hourglass Network~\cite{newell2016stacked}, FPN~\cite{yang2017learning}, CPN~\cite{chen2018cascaded}, SimpleBaseline~\cite{xiao2018simple}, HRNet~\cite{sun2019hrnet}, RSN~\cite{cai2020learning}, even automated architectures~\cite{yang2019pose,gong2020autopose,mcnally2020evopose2d, cheng2020scalenas, zhang2020efficientpose}. But as the architecture is becoming more complex, it is more challenging but imperative than ever to seek the interpretability of human pose estimation models. In contrast, our model can estimate human pose in an efficient and explicit way. 

\subsection{Explainability}

Explainability means a better understanding for human of how the model makes predictions. As surveyed by~\cite{samek2019explainable}, many works define the goal for explanation is to determine what inputs are the most relevant to the prediction, which is also \emph{the goal we seek in this paper}. ~\cite{erhan2009visualizing, li2015heterogeneous} perform gradient descent in the input space to find out what input patterns can maximize a given unit. ~\cite{simonyan2013deep, fang2017rmpe} further consider generating the image-specific class saliency maps. ~\cite{zeiler2014visualizing} uses DeConvNet to generate feature activities to show what convolutional layers have learned. Some pose estimation methods ~\cite{li2015heterogeneous,zhang2018occluded} visualize the feature maps by choosing specific neurons or channels but the results fail to reveal the spatial relationship between parts. ~\cite{Tang_2019_CVPR} estimates the probability distributions and mutual information between keypoints, yet only revealing the statistic information rather than image-specific explanations. There are also works like Network Dissection~\cite{bau2017network}, Feature Visualization~\cite{olah2017feature}, Excitation Backprop~\cite{zhang2016top}, LRP attribution method~\cite{bach2015pixel}, CAM~\cite{zhou2016learning}, and Grad-CAM~\cite{selvaraju2017grad}, which aim to explain the prediction of CNN classifier or visualize the saliency area significantly affecting the class. Different from most prior works, we aim to reveal the fine-grained spatial dependencies between body joints variables in the structural skeleton. And our model can directly exploit the attention patterns to holistically explain its predictions without the help of external tools. We also notice a recent paper~\cite{chefer2020transformer} that develops LRP-based~\cite{bach2015pixel} method to compute relevance to explain the predictions of Transformer. It takes ViT model~\cite{dosovitskiy2020an} to visualize class-specific relevance map, showing reasonable results. Unlike their goal, we focus on revealing what clues contribute to visual keypoint localizations, and the attentions in our model provide clear evidence for the predictions.

It is worth noting that there are some works, such as CoordConv~\cite{liu2018an} and Zero Padding~\cite{islam2020how}, to explain how the neural network predicts the positions and stores the position information by designing proxy tasks. We also conduct experiments to investigate the importance of position embedding for predicting the locations and its generalization on unseen input scales.

\subsection{Transformer}

Transformer was proposed by Vaswani \emph{et al.}~\cite{vaswani2017attention} for neural machine translation (NMT) task~\cite{sutskever2014sequence}. Large Transformer-based models like BERT~\cite{devlin2018bert}, GPT-2~\cite{radford2019language} are often pre-trained on large amounts of data and then fine-tuned for smaller datasets. Recently, Vision Transformer or attention-augmented layers have merged as new choices for vision tasks such as \cite{parmar2018image, ramachandran2019stand, bello2019attention, dosovitskiy2020an,touvron2020deit, carion2020detr, chen2020pre, dai2020up, zhu2020deformable, wang2020end}. DETR~\cite{carion2020detr} directly predicts a set of object instances by introducing object queries. ViT~\cite{dosovitskiy2020an} is to pre-train a pure Transformer on large data and then fine-tuned on ImageNet for image classification. DeiT~\cite{touvron2020deit} introduces a distillation token to learn knowledge from a teacher. There are also works~\cite{epipolartransformers2020he, handtransformer:huang2020hand, metro:lin2021end} applying Transformers to 3D pose estimation. \cite{epipolartransformers2020he} fuses features from multi-view images by attention mechanism.~\cite{handtransformer:huang2020hand, metro:lin2021end} output 1D sequences composed of joint/vertex coordinates of pose. Unlike them, we use Transformer to predict the 2D heatmaps represented with spatial distributions of keypoints for 2D human pose estimation problem.

\section{Method}
Our goal is to build a model that can explicitly capture global dependencies between human body parts.
We first describe the model architecture. Then we show how it exploits self-attention to capture global interactions and establish a connection between our method and the principle of Activation Maximization.
\subsection{Architecture}
\begin{figure*}[t]
	\begin{center}
		\includegraphics[width=1\linewidth]{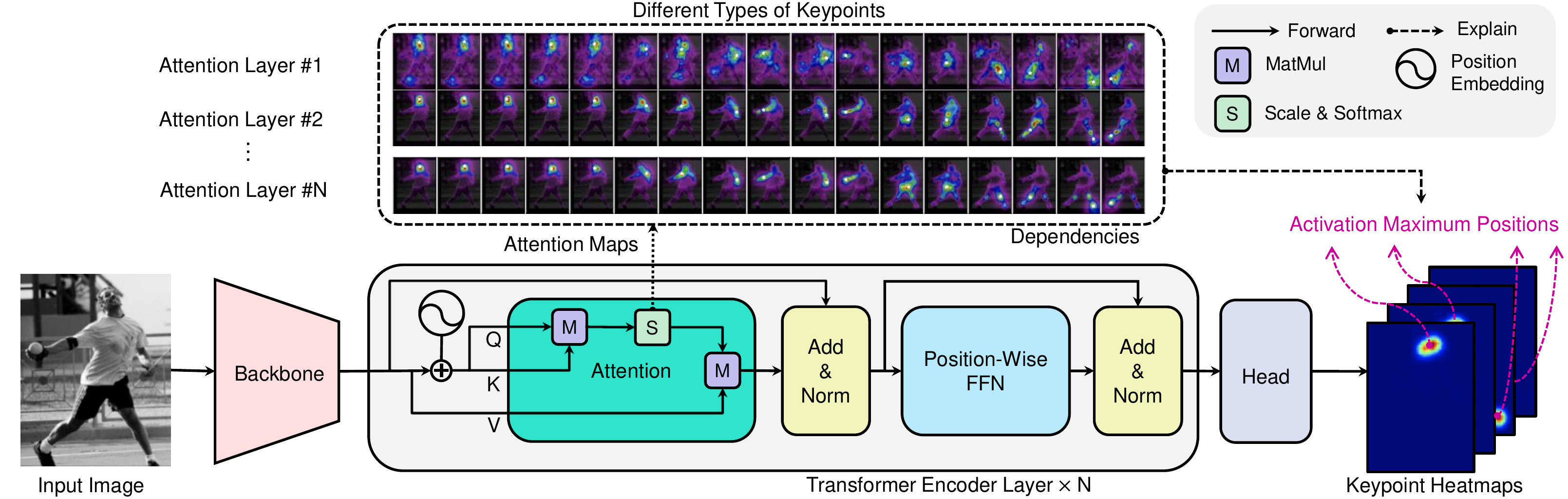}
	\end{center}
	\caption{The architecture. Firstly, the feature maps are extracted by a CNN backbone and flattened into a sequence. Next, the Transformer encode layers iteratively capture dependencies from the sequences by query-key-value attention. Then, a simple head is used to predict the keypoints heatmaps. The attention map in Transformer can reveal what dependencies (regions or joints) significantly contribute to the activation maximum positions in the predicted keypoint heatmaps.}\vspace*{-0.1in}
	\label{architecture}
\end{figure*}

As illustrated in Fig.~\ref{architecture}, TransPose model consists of three components: a CNN backbone to extract low-level image feature; a Transformer Encoder to capture long-range spatial interactions between feature vectors across the locations; a head to predict the keypoints heatmaps.

{\bf Backbone.} Many common CNNs can be taken as the backbone. For better comparisons, we choose two typical CNN architectures: ResNet~\cite{he2016deep} and HRNet~\cite{sun2019hrnet}. We only retain the initial several parts of the original ImageNet pretrained CNNs to extract feature from images. We name them ResNet-S and HRNet-S, the parameters numbers of which are only about 5.5\% and 25\% of the original CNNs. 

{\bf Transformer.} We follow the standard Transformer architecture~\cite{vaswani2017attention} as closely as possible. And only the Encoder is employed, as we believe that the pure heatmaps prediction task is simply an encoding task, which compresses the original image information into a compact position representation of keypoints. Given an input image $I\in\mathbb{R}^{ 3\times H_I \times W_I}$, we assume that the CNN backbone outputs a 2D spatial structure image feature $\mathbf{X}_f\in \mathbb{R}^{ d\times H \times W}$ whose feature dimension has been transformed to $d$ by a 1$\times$1 convolution. Then, the image feature map is flattened into a sequence $\mathbf{X}\in \mathbb{R}^{ L\times d}$, \emph{i.e.}, $L$ $d$-dimensional feature vectors where $L=H\times W$. It goes through $N$ attention layers and feed-forward networks (FFNs). 

{\bf Head.} A head is attached to Transformer Encoder output $\mathbf{E}\in\mathbb{R}^{L\times d}$ to predict $K$ types of keypoints heatmaps $P\in\mathbb{R}^{ K\times H^* \times W^*}$ where $H^*,W^*=H_I/4,W_I/4$ by default. We firstly reshape $\mathbf{E}$ back to $\mathbb{R}^{ d\times H \times W}$ shape. Then we mainly use a 1$\times$1 convolution to reduce the channel dimension of $\mathbf{E}$ from $d$ to $K$. If $H,W$ are not equal $H^*,W^*$, an additional bilinear interpolation or a 4$\times$4 transposed convolution is used to do upsampling before 1$\times$1 convolution. Note, a 1$\times$1 convolution is completely equivalent to a \emph{position-wise linear} transformation layer. 

\subsection{Resolution Settings.}

Due to that the computational complexity of per self-attention layer is $O\left((HW)^{2} \cdot d\right)$, we restrict the attention layers to operate at a resolution with $r\times$ downsampling rate w.r.t. the original input, \emph{i.e.}, $H,W =H_I /r,W_I /r$. In the common human pose estimation architectures~\cite{wei2016convolutional, newell2016stacked, xiao2018simple, sun2019hrnet}, $32\times$ downsampling is usually adopted as a standard setting to obtain a very low-resolution map containing global information. In contrast, we adopt $r=8$ and $r=4$ setting for ResNet-S and HRNet-S, which are beneficial to the trade-off between the memory footprint for attention layers and the loss in detailed information. As a result, our model directly captures long-range interactions at a higher resolution, while preserving the fine-grained local feature information.

\begin{table*}\scriptsize 
	\begin{center}
		\renewcommand{\arraystretch}{0.85}
		\setlength{\tabcolsep}{2.0mm}
		\begin{tabular}{l|ccc|cccc|c}
			\toprule
			\textbf{Model Name} & \textbf{Backbone} &\textbf{Downsampling for Attention} & \textbf{Upsampling} &\#\textbf{Layers} &\textbf{Heads}& \textbf{d }& \textbf{h }& \#\textbf{Params} \\
			\midrule
			
			TransPose-R-A3* &  ResNet-Small*  & 1/8 &  Bilinear Interpolation &3 & 8 & 256 & 512 &5.0M \\
			TransPose-R-A3 &  ResNet-Small  & 1/8 &Deconvolution &3 & 8 & 256 & 1024 &5.2M \\
			TransPose-R-A4 &  ResNet-Small  & 1/8 &Deconvolution &4 & 8 & 256 & 1024 &6.0M \\
			
			TransPose-H-S &  HRNet-Small-W32  & 1/4 & None& 4& 1 & 64 & 128    & 8.0M \\
			TransPose-H-A4 &  HRNet-Small-W48   & 1/4 &None& 4& 1 & 96 & 192    & 17.3M \\
		
			TransPose-H-A6 &  HRNet-Small-W48   & 1/4 &None& 6& 1 & 96 & 192    & 17.5M \\	
			\bottomrule
		\end{tabular}
	\end{center}
	\caption{Architecture configurations for different TransPose models. More details about the backbones are described in supplementary.}
	\label{architecture configurations}
\end{table*}

\begin{table*}\footnotesize
	\begin{center}
		\setlength{\tabcolsep}{4.5mm}
		\renewcommand{\arraystretch}{0.9}
		\begin{tabular}{l|c|ll|l|c|l}
			\toprule[0.11em]
			\textbf{Method} &\textbf{Input Size} &\textbf{AP} & \textbf{AR}& \#\textbf{Params} &\textbf{FLOPs} & \textbf{FPS} \\
			\midrule
		
			SimpleBaseline-Res50~\cite{xiao2018simple} &256$\times$192 &70.4& 76.3 &34.0M &8.9G & 114\\
			SimpleBaseline-Res101~\cite{xiao2018simple} &256$\times$192 &71.4& 76.3 &53.0M &12.4G & 92\\
			SimpleBaseline-Res152~\cite{xiao2018simple} &256$\times$192 &72.0& 77.8 &68.6M &35.3G & 62\\
			\hline
			TransPose-R-A3* & 256$\times$192 & 71.5 & 76.9&5.0M ({\color{black}$\downarrow$85\%}) &5.4G & 137 ({\color{black}$\uparrow$20\%})\\
			TransPose-R-A3 & 256$\times$192 & 71.7 & 77.1&5.2M ({\color{black}$\downarrow$85\%}) &8.0G & 141 ({\color{black}$\uparrow$23\%})\\
			
			TransPose-R-A4 & 256$\times$192 & \textbf{72.6} & \textbf{78.0}& 6.0M ({\color{black}$\downarrow$82\%}) &8.9G &138 ({\color{black}$\uparrow$21\%})\\
			
			\midrule
			HRNet-W32~\cite{sun2019hrnet} &256$\times$192 &74.4& 79.8 &28.5M &7.2G & 28\\
			
			HRNet-W48~\cite{sun2019hrnet} &256$\times$192 &75.1& 80.4 &63.6M &14.6G & 27\\
			
			\hline
			
			TransPose-H-S &256$\times$192 & 74.2 & 78.0& 8.0M ({\color{black}$\downarrow$72\%})& 10.2G & 45 ({\color{black}$\uparrow$61\%})\\

			TransPose-H-A4 &256$\times$192 & 75.3 & 80.3& 17.3M ({\color{black}$\downarrow$73\%})& 17.5G & 41 ({\color{black}$\uparrow$52\%})\\
		
			TransPose-H-A6 &256$\times$192 & \textbf{75.8} & \textbf{80.8}& 17.5M ($\downarrow$73\%)& 21.8G & 38 ({\color{black}$\uparrow$41\%})\\
			\bottomrule[0.1em]
		\end{tabular}
	\end{center}
	\caption{Results on COCO validation set, all provided with the same detected human boxes. TransPose-R-* and TransPose-H-* achieve competitive results to SimpleBaseline and HRNet, with fewer parameters and faster speeds. The reported FLOPs of SimpleBaseline and HRNet only include the convolution and linear layers.}\vspace*{-0.1in} 
	\label{state-of-the-art}
\end{table*}

\subsection{Attentions are the Dependencies of Localized Keypoints}

\label{attention mechanism}
{\bf Self-Attention mechanism.} The core mechanism of Transformer~\cite{vaswani2017attention} is multi-head self-attention. It first projects an input sequence $\mathbf{X} \in \mathbb{R}^{ L\times d}$ into queries $\mathbf{Q}\in \mathbb{R}^{ L\times d}$, keys $\mathbf{K}\in \mathbb{R}^{ L\times d}$ and values $\mathbf{V}\in \mathbb{R}^{ L\times d}$ by three matrices $\mathbf{W}_q,\mathbf{W}_k,\mathbf{W}_v \in \mathbb{R}^{d\times d}$. Then, the attention scores matrix\footnote{Here we consider single-head self attention. For multi-head self-attention, the attention matrix is the average of attention maps in all heads.} $\mathbf{A}\in\mathbb{R}^{ N\times N}$ is computed by: 
\begin{equation}
\mathbf{A}=\operatorname{softmax}\left(\frac{\mathbf{Q} \mathbf{K}^\top}{\sqrt{d}}\right).
\end{equation}
Each query $\boldsymbol{q}_i\in\mathbb{R}^d$ of the token $\boldsymbol{x}_i\in \mathbb{R}^d$ (i.e., feature vector at location $i$) computes similarities with all the keys to achieve a weight vector $\mathbf{w}_i=\mathbf{A}_{i,:} \in \mathbb{R}^{1\times L}$, which determines how much dependency is needed from each token in the previous sequence. Then an increment is achieved by a linear sum of all elements in Value matrix $\mathbf{V}$ with the corresponding weight in $\mathbf{w}_i$ and added to $\boldsymbol{x}_i$. By doing this, the attention maps can be seen as \emph{dynamic weights} that determined by specific image content, reweighting the information flow in the forward propagation. 

Self-attention captures and reveals how much contribution the predictions aggregate from each image location. Such contributions from different image locations can be reflected by the gradient~\cite{simonyan2013deep, bach2015pixel, selvaraju2017grad}. Therefore, we concretely analyze how $\boldsymbol{x}_j$ at image/sequence location $j$ affects the activation $\boldsymbol{h}_i$ at location $i$ the predicted keypoint heatmaps, by computing the derivative of $\boldsymbol{h}_i\in \mathbb{R}^{K}$ ($K$ types of keypoints) w.r.t the $\boldsymbol{x}_j$ at location $j$ of the input sequence of the last attention layer. And we further assume $G:=\frac{\partial \boldsymbol{h}_i}{\partial \boldsymbol{x}_j}$ as a function w.r.t. a given attention score $\mathbf{A}_{i,j}$. We obtain:
\begin{equation}\label{grad}
\begin{aligned}
G\left(\mathbf{A}_{i,j} \right) 
&\approx\mathbf{A}_{i,j}\cdot\mathbf{W}_f
\cdot\mathbf{W}_v^\top+\mathbf{W}_f=\mathbf{A}_{i,j}\cdot\mathbf{K} + \mathbf{B}
\end{aligned}
\end{equation}
where $\mathbf{K},\mathbf{B}\in\mathbb{R}^{K\times d}$ are \emph{static weights} (fixed when inferring) and shared across all image locations. The derivations of Eq.~\ref{grad} are shown in supplementary. We can see that the function $G$ is approximately linear with $\mathbf{A}_{i,j}$, \emph{i.e.}, the degrees of contribution to the prediction $\boldsymbol{h}_i$ directly depend on its attention scores at image locations.  

Especially, the last attention layer acts as \emph{an aggregator}, which collects contributions from all image locations according to attentions and forms the maximum activations in the predicted keypoint heatmaps. Although the layers in FFN and head cannot be ignored, they are \emph{position-wise}, which means they approximately linearly transform the contributions from all locations by the same transformation without changing their relative proportions.

\label{paper::grad}\label{paper::definition}
{\bf The activation maximum positions are the keypoints' locations.} The interpretability of Activation Maximization (AM)~\cite{erhan2009visualizing,simonyan2013deep} lies in: the input region which can maximize a given neuron activation can explain what this activated neuron is looking for. 

In this task, the learning target of TransPose is to expect the neuron activation $h_{i^*}$ at location $i^*$ of the heatmap to be maximally activated where $i^*$ represents the groundtruth location of a keypoint:
\begin{equation}
\quad\theta^*=\arg\max_{\theta}h_{i^*}(\theta,I).
\end{equation}
Assuming the model has been optimized with parameters $\theta^*$ and it predicts the location of a particular keypoint as $i$ (maximum position in a heatmap), why the model predicts such prediction can be explained by the fact that those locations $\mathbf{J}$, whose element $j$ has higher attention score ($\geq\delta$) with $i$, are the dependencies that significantly contribute to the prediction. The dependencies can be found by:
\begin{equation}
\mathbf{J}=\left\lbrace j| \mathbf{A}_{i,j}\left( \theta^*,I\right) \geq\delta\right\rbrace,
\end{equation}
where $\mathbf{A}\in\mathbb{R}^{ L\times L}$ is the attention map of the last attention layer and also a function w.r.t $\theta^*$ and $I$, \emph{i.e.}, $\mathbf{A}=\mathbf{A}\left( \theta^*,I\right)$. Given an image $I$ and a query location $i$, $\mathbf{A}_{i,:}$ can reveal what dependencies a predicted location $i$ highly relies on, we define it \textbf{\emph{dependency area}}. $\mathbf{A}_{:,j}$ can reveal what area a location $j$ mostly affects, we define it \emph{affected area}.

For the traditional CNN-based methods, they also use heatmap activations as the keypoint locations, but one cannot directly find the explainable patterns for the predictions due to the deepness and highly non-linearity of deep CNNs. The AM-based methods~\cite{erhan2009visualizing, li2015heterogeneous, zeiler2014visualizing, simonyan2013deep} may provide insights while they require extra optimization costs to learn explainable patterns the convolutional kernels prefer to look for. Different from them, we extend AM to heatmap-based localization via Transformer, and we do not need extra optimization costs because the optimization has been implicitly accomplished in our training, i.e., $\mathbf{A}=\mathbf{A}\left( \theta^*,I\right)$.  The defined \textbf{\emph{dependency area}} is the pattern we seek, which can show image-specific and keypoint-specific dependencies. 

\section{Experiments}

{\bf Dataset.} We evaluate our models on COCO~\cite{lin2014microsoft} and MPII~\cite{andriluka2014} datasets. COCO contains 200k images in the wild and 250k person instances. Train2017 consists of 57k images and 150k person instances. Val2017 set contains 5k images and test-dev2017 consists of 20k images. In Sec~\ref{transfer-mpii}, we show the experiments on MPII~\cite{andriluka2014}. And we adopt the standard evaluation metrics of these benchmarks.

{\bf Technical details.} We follow the top-down human pose estimation paradigm. The training samples are the cropped images with single person. We resize all input images into $256\times192$ resolution. We use the same training strategies, data augmentation and person detected results as~\cite{sun2019hrnet}. We also adopt the coordinate decoding strategy proposed by~\cite{zhang2020distribution} to reduce the quantisation error when decoding from downscaled heatmaps. The feed forward layers are trained with 0.1 dropout and ReLU activate function. Next, we name the models based on ResNet-S and HRNet-S \emph{TransPose-R} and \emph{TransPose-H}, abbreviated as \emph{\textbf{TP-R}} and \emph{\textbf{TP-H}}. The architecture details are reported in Tab.~\ref{architecture configurations}. We use Adam optimizer for all models. Training epochs are 230 for TP-R and 240 for TP-H. The cosine annealing learning rate decay is used. The learning rates for TP-R-A4 and TP-H-A6 models decay from 0.0001 to 0.00001, we recommend using such a schedule for all models. Considering the compatibility with backbone and the memory consumption, we adjust the hyperparameters of Transformer encoder to make the model capacity not very large. In addition, we use 2D sine position embedding as the default position embedding. We describe it in the supplementary.

\subsection{Results on COCO keypoint detection task}
 We compare TransPose with SimpleBaseline, HRNet, and DARK~\cite{zhang2020distribution}. Specially, we trained the DARK-Res50 on our machines according to the official code with TransPose-R-A4's data augmentation, we achieve 72.0AP; when using the totally same data augmentation and long training schedule of TransPose-R-A4 for it, we obtain 72.1AP (+0.1 AP). The other results showed in Tab.~\ref{state-of-the-art} come from the papers. We test all models on a single NVIDIA 2080Ti GPU with the same experimental conditions to compute the average FPS. Under the input resolution -- 256$\times$192, TransPose-R-A4 and TransPose-H-A6 have obviously overperformed SimpleBaseline-Res152 (+0.6AP)~\cite{xiao2018simple}, HRNet-W48 (+0.7AP)~\cite{sun2019hrnet} and DARK-HRNet~\cite{zhang2020distribution} (+0.2AP), with significantly fewer model parameters and faster speeds. Tab.~\ref{coco-test} shows the results on COCO test set.

\begin{table}\scriptsize 
	
	\centering
	\label{table:coco_test_dev}
	\renewcommand{\arraystretch}{0.9}
	\setlength{\tabcolsep}{0.2mm}
	\begin{tabular}{c|c|ccc|ccccc}
		\toprule[0.1em]
		\textbf{Method} & \textbf{Input size} & \#\textbf{Params} & \textbf{FLOPs}& \textbf{FPS}&
		$\textbf{AP}$ & $\textbf{AP}_{\text{0.5}}$ & $\textbf{AP}_{\text{0.75}}$ & $\textbf{AP}_{\text{M}}$ & $\textbf{AP}_{\text{L}}$ \\
		\midrule
		
		G-RMI~\cite{papandreou2017towards}  & 353$\times$257 & 42.6M  & 57G&- &64.9 & 85.5&71.3&62.3&70.0\\
		Integral~\cite{sun2018integral} & 256$\times$256 &45.0M &11.0G&- &67.8 & 88.2&74.8&63.9&74.0\\
		
		CPN~\cite{chen2018cascaded}& 384$\times$288&58.8M&29.2G&-
		& 72.1 & 91.4&80.0&68.7&77.2\\
		RMPE~\cite{fang2017rmpe} & 320$\times$256 &28.1M &26.7G&-
		&72.3 & 89.2&79.1&68.0&78.6\\
		
		SimpleBaseline~\cite{xiao2018simple} &384$\times$288  &68.6M & 35.6G&-
		&73.7 & 91.9&81.1&70.3&80.0\\
		HRNet-W32~\cite{sun2019hrnet} & 384$\times$288 &28.5M& 16.0G&26&74.9&92.5&82.8&71.3&80.9\\
		HRNet-W48~\cite{sun2019hrnet} & 256$\times$192 &63.6M&14.6G&27& 74.2&92.4&82.4&70.9&79.7\\
		HRNet-W48~\cite{sun2019hrnet} & 384$\times$288 &63.6M&32.9G&25& 75.5&92.5&83.3&71.9&81.5\\
		DarkPose~\cite{zhang2020distribution} & 384$\times$288 &63.6M&32.9G&25& 76.2&92.5&83.6&72.5&82.4\\
		\midrule
		\textbf{TransPose-H-S} & 256$\times$192 &8.0M&10.2G&45& 73.4&91.6&81.1&70.1&79.3\\
		\textbf{TransPose-H-A4} & 256$\times$192 &17.3M&17.5G&41& 74.7&91.9&82.2&71.4&80.7\\
		
		\textbf{TransPose-H-A6} & 256$\times$192 &17.5M&21.8G&38& 75.0&92.2&82.3&71.3&81.1\\
		\bottomrule[0.1em]

	\end{tabular}
	\caption{Comparisons with state-of-the-art CNN-based models on COCO test-dev set. Tested on smaller input resolution 256$\times$192 , our models achieve comparable performances with the others.}\vspace*{-0.01in}
	\label{coco-test}
\end{table}

\label{position embedding}
\begin{table}[h]\footnotesize
	\begin{center}
		\setlength{\tabcolsep}{2mm}
		\renewcommand{\arraystretch}{0.6}
		\begin{tabular}{c|ccc}
			\toprule[0.1em]
			\textbf{Position Embedding} & \#\textbf{Params} & \textbf{FLOPs} &\textbf{AP}   \\
			
			\midrule
			\ding{55}& 4.999M & 7.975G& 70.4  \\ 
			Learnable & 5.195M & 7.976G& 70.9 \\ 
			2D Sine (Fixed)& 5.195M&7.976G&71.7 \\ 
			\bottomrule[0.1em]
		\end{tabular}
	\end{center}
	\caption{Results for different position embedding schemes for TransPose models. The input size is $256\times192$.}\vspace*{-0.01in}
	\label{position embedding tab}
\end{table}

\subsection{Transfer to MPII benchmark}
\label{transfer-mpii}
Typical pose estimation methods often separately train and evaluate their models on COCO and MPII~\cite{andriluka2014}. Motivated by the success of pre-training in NLP and recent ViT~\cite{dosovitskiy2020an}, we try to transfer our pre-trained models to MPII. We replace the final layer of the pre-trained TransPose model with a uniform-initialized $d \times 16$ linear layer for MPII. When fine-tuning, the learning rates for the pre-trained and final layers are 1$e$-5 and 1$e$-4 with decay. 
\begin{figure}
	\hspace{-0.5em}	\vspace{-0.02in}
	\includegraphics[width=1\linewidth]{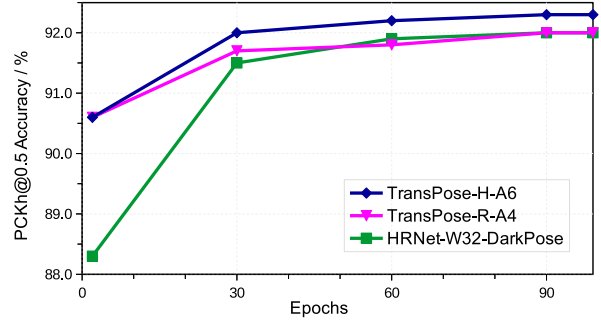}
	\vspace*{-0.02in}
	\caption{Performances on validation set when fine-tuning models (listed in Tab.~\ref{table::mpii-comparisons}) with different epochs on MPII training set.} \vspace{-0.01in}
	\label{figure::mpii-finetune}
\end{figure}

\begin{table}\scriptsize 
	\centering
	\renewcommand{\arraystretch}{0.9}
	\setlength{\tabcolsep}{0.9mm}
	\begin{tabular}{c|ccllc}
		\toprule[0.1em]
		\textbf{Models} &\textbf{ Strategy} &\textbf{ Epochs} &\textbf{ Mean@0.5}&\textbf{ Mean@0.1}&\#\textbf{Params}\\
		\midrule
		\multirow{2}{*}{DARK-HRNet~\cite{zhang2020distribution}} & $\circlearrowleft$ & 210 & 90.6 & 42.0 & 28.5M\\
	
		& $\Rightarrow$ & 100 & 92.0 (+1.4) & 43.6 (+1.6) & 28.5M\\
		\midrule
		\multirow{2}{*}{TransPose-R-A4} &$\circlearrowleft$ & 230 & 89.3 & 38.6 & 6.0M \\
		& $\Rightarrow\uparrow$ & 100 & 92.0 (\textbf{+2.7}) & 44.1 (\textbf{+5.5}) & 6.0M\\
		\midrule
		\multirow{2}{*}{TransPose-H-A6} & $\circlearrowleft$ & 230 & 90.3 & 41.6 & 17.5M\\
		& $\Rightarrow$ & 100 & \textbf{92.3} (+2.0) & \textbf{44.4} (+2.8) & 17.5M\\
		\bottomrule[0.1em]
	\end{tabular}
	\caption{Fine-tuning and full-training performances on MPII validation set. $\circlearrowleft$ means full-training on MPII without COCO pre-training. $\Rightarrow$ means transferring the pretrained model and fine-tuning on MPII; adding $\uparrow$ means fine-tuning MPII on input resolution 384$\times$384 otherwise 256$\times$256.}\vspace*{-0.1in}
	\label{table::mpii-comparisons}
\end{table}

For comparisons, we fine-tune the pre-trained DARK-HRNet on MPII with the same settings, and train these models on MPII by standard full-training settings. As shown in Tab.~\ref{table::mpii-comparisons} and Fig.~\ref{figure::mpii-finetune}, the results are interesting: even with longer full-training epochs, models perform worse than the fine-tuned ones; even with large model capacity (28.5M), the improvement (+1.4 AP) brought by pre-training DARK-HRNet is smaller than pre-training TransPose (+2.0 AP). With 256$\times$256 input resolution and fine-tuning on MPII train and val sets, the best result on MPII test set yielded by TransPose-H-A6 is 93.5\% accuracy, as shown in Fig.~\ref{table::mpii-test}. These results show that pre-training and fine-tuning could significantly reduce training costs and improve the performances, particularly for the pre-trained TransPose models.

\begin{table}\scriptsize 
	\centering
	\renewcommand{\arraystretch}{0.9}
	\setlength{\tabcolsep}{1mm}
	\begin{tabular}{c|ccc}
		\toprule[0.1em]
		\textbf{Method} & \textbf{Input size} & \textbf{Training Data} & \textbf{Mean@0.5}\\
		\midrule
		Belagiannis \& Zisserman, FG'17 ~\cite{belagiannis2017recurrent}& 248$\times$248& COCO+MPII$\dagger$ &88.1 \\
		Su et al., arXiv'19~\cite{su2019cascade} & 384$\times$384 & HSSK+MPII$\ddagger$ & 93.9 \\
		Bulat et al., FG'20~\cite{bulat2020toward} & 256$\times$256 & HSSK+MPII$\ddagger$ & 94.1 \\
		Bin et al., ECCV'20~\cite{bin2020adversarial} & 384$\times$384 & HSSK+MPII$\ddagger$ & 94.1 \\
		
		\midrule
		Ours (TransPose-H-A6) & 256$\times$256 & COCO+MPII$\dagger$ & 93.5 \\
		\bottomrule[0.1em]
	\end{tabular}
	\caption{Results on MPII benchmark test set. $\dagger$ means pre-training on COCO dataset and fine-tuning on MPII dataset. $\ddagger$ means training both on MPII and HSSK datasets.}\vspace*{-0.1in}
	\label{table::mpii-test}
\end{table}

 {\bf Discussion.} The pre-training and fine-tuning for Transformer-based models have shown favorable results in NLP~\cite{devlin2018bert,radford2019language} and recent vision models~\cite{dosovitskiy2020an,chen2020pre,dai2020up}. 
 Our initial results on MPII also suggest that training Transformer-based models on large-scale pose-related data may be a promising way to learn powerful and robust representation for human pose estimation and its downstream tasks.

\subsection{Ablations}

{\bf The importance of position embedding.} Without position embedding, the 2D spatial structure information loses in Transformer. To explore its importance, we conduct experiments on TransPose-R-A3 models with three position embedding strategies: 2D sine position embedding, learnable position embedding, and w/o position embedding. As expected, the models with position embedding perform better, particularly for 2D sine position embedding, as shown in Tab.~\ref{position embedding tab}. But interestingly, TransPose w/o any position embedding only loses 1.3 AP, which suggests that 2D-structure becomes less important. See more details in supplementary.

{\bf Scaling the Size of Transformer Encoder.} We study how performance scales with the size of Transformer Encoder, as shown in Tab.~\ref{layers}. For TransPose-R models, with the number of layers increasing to 6, the performance improvements gradually tend to saturate or degenerate. But we have not observed such a phenomenon on TransPose-H models. Scaling the Transformer obviously improves the performance of TransPose-H. 

\begin{figure}
	\begin{center}
		\includegraphics[width=1\linewidth]{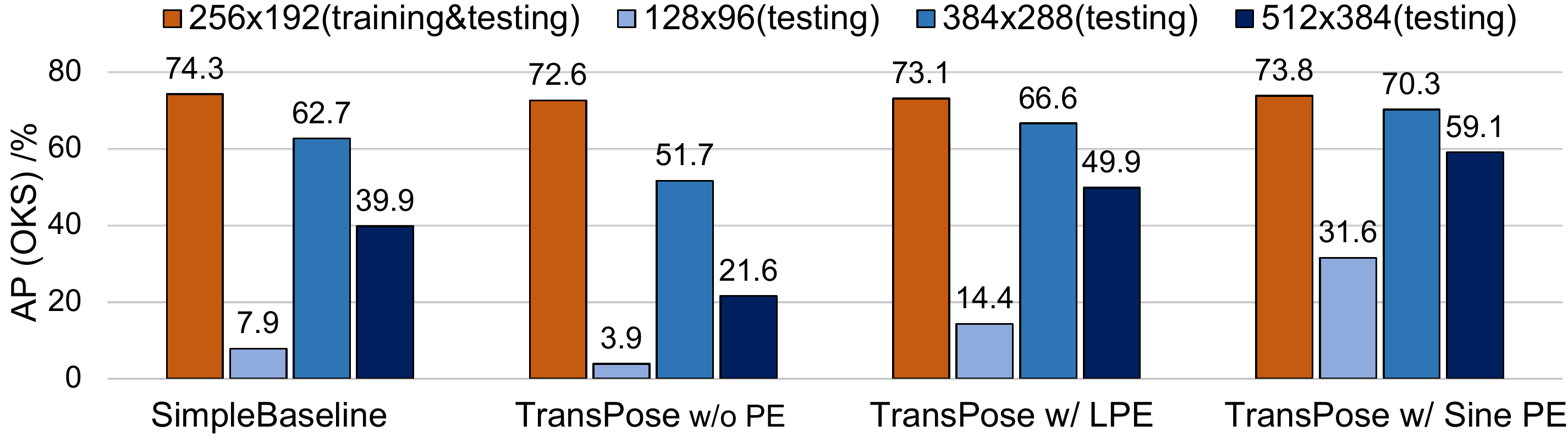}
	\end{center}\vspace*{-0.15in}
	\caption{Performances on unseen input resolutions. TransPose models w/ Position Embedding generalize better.} \vspace{-0.1in}
	\label{generalization_fig}
\end{figure}

\label{unseen}
{\bf Position embedding helps to generalize better on unseen input resolutions.} The top-down paradigm scales all the cropped images to a fixed size. But for some cases even with a fixed input size or the bottom-up paradigm, the body size in the input varies; the robustness to different scales becomes important. So we design an extreme experiment to test the generalization: we test SimpleBaseline-ResN50-Dark and TransPose-R-A3 models on unseen 128$\times$96, 384$\times$288, 512$\times$388 input resolutions, all of which only have been trained with 256$\times$192 size. Interestingly, the results in Fig.~\ref{generalization_fig} demonstrate that SimpleBaseline and TransPose-R w/o position embedding have obvious performance collapses on unseen resolutions, particularly on 128$\times$96; but TransPose-R with learnable or 2D Sine position embedding have significantly better generalization, especially for 2D Sine position embedding. 

{\bf Discussion.} For the input resolution, we mainly trained our models on 256$\times$192 size, thus 768 and 3072 sequence lengths for Transformers in TP-R and TP-H models. Higher input resolutions such as 384$\times$288 for our current models will bring prohibitively expensive computational costs in self-attention layers due to the quadratic complexity.



\begin{table}\scriptsize
	\centering
	\renewcommand{\arraystretch}{0.6}
	\setlength{\tabcolsep}{1mm}
	\begin{tabular}{ccccccccc}
		\toprule[0.1em]
		\textbf{Model} & \#\textbf{Layers} & $d$ & $h$ & \#\textbf{Params} & \textbf{FLOPs}& \textbf{FPS}& \textbf{AP} &\textbf{ AR}\\
		\midrule
		\multirow{5}{*}{TransPose-R}&\cellcolor[gray]{0.99}2& 256& 1024& 4.4M & 7.0G & 174 & \cellcolor[gray]{0.99}69.6 & \cellcolor[gray]{0.99}75.0\\
		&\cellcolor[gray]{0.96}3& 256& 1024 & 5.2M & 8.0G & 141 & \cellcolor[gray]{0.96}71.7 & \cellcolor[gray]{0.96}77.1 \\
		&\cellcolor[gray]{0.93}4& 256& 1024 & 6.0M & 8.9G & 138 & \cellcolor[gray]{0.87}72.6 & \cellcolor[gray]{0.87}78.0 \\
		&\cellcolor[gray]{0.90}5& 256& 1024 & 6.8M & 9.9G & 126 & \cellcolor[gray]{0.93}72.2 & \cellcolor[gray]{0.93}77.6 \\
		&\cellcolor[gray]{0.87}6& 256& 1024 & 7.6M & 10.8G & 109 & \cellcolor[gray]{0.93}72.2 & \cellcolor[gray]{0.93}77.5 \\
		\midrule
		\multirow{5}{*}{TransPose-H}&4& \cellcolor{cyan!5}64& \cellcolor{cyan!5}128 & 17.0M & 14.6G & - &  \cellcolor{cyan!5}75.1 &  \cellcolor{cyan!5}80.1 \\
		&4&  \cellcolor{cyan!10}192&  \cellcolor{cyan!10}384 & 18.5M & 27.0G & - &  \cellcolor{cyan!10}75.4 &  \cellcolor{cyan!10}80.5 \\
		
		&\cellcolor{blue!5}4& 96& 192 & 17.3M & 17.5G & 41 & \cellcolor{blue!5}75.3 & \cellcolor{blue!5}80.3 \\
		&\cellcolor{blue!10}5& 96& 192 & 17.4M & 19.7G & 40 & \cellcolor{blue!10}75.6 & \cellcolor{blue!10}80.6 \\
		&\cellcolor{blue!15}6& 96& 192 & 17.5M & 21.8G & 38 & \cellcolor{blue!15}75.8 & \cellcolor{blue!15}80.8 \\
		\bottomrule[0.1em]
	\end{tabular}
	\caption{Ablation study on the size of Transformer Encoder. \#Layers, $d$ and $h$ are the number of encoder layers, the dimensions $d$, and the number of hidden units of FFN. }\vspace*{-0.01in}
	\label{layers}
\end{table}

\begin{figure*}[t]
	\centering
	\hspace{-0.5em}
	\subfigure[\textbf{TP-R-A4:} predicted keypoints and their \emph{dependency areas} for input \textbf{A}.]{
		\includegraphics[width=0.48\linewidth]{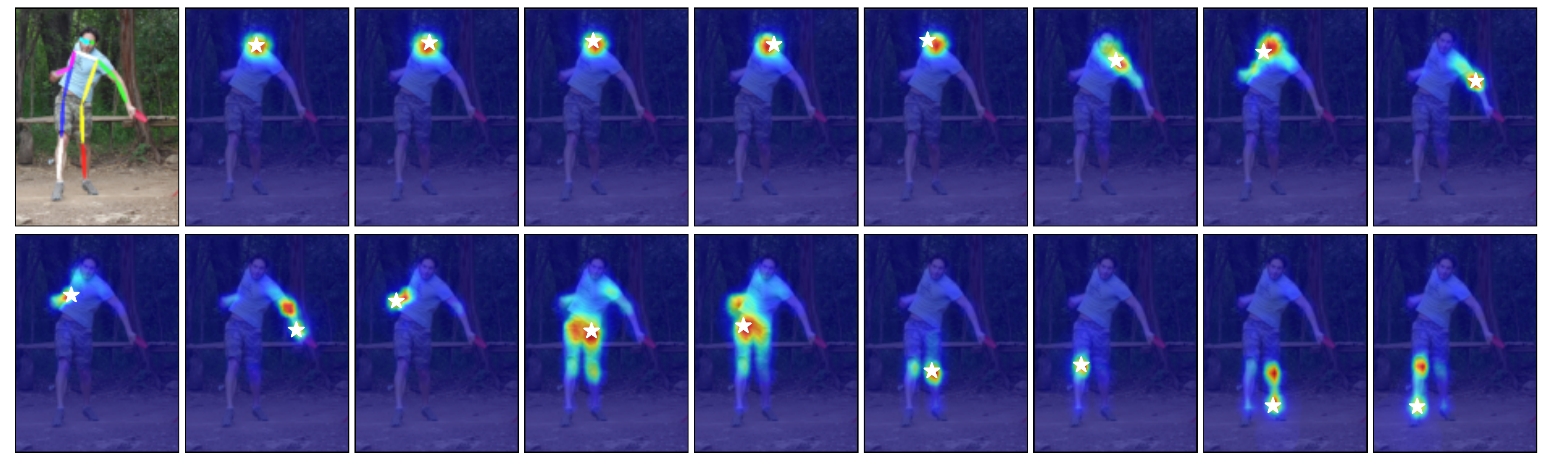}
		\label{tpr-a}
	}\hspace{1.em}\vspace{-0.01in}
	\subfigure[\textbf{TP-H-A4:} predicted keypoints and their \emph{dependency areas} for input \textbf{A}.]{
		\includegraphics[width=0.48\linewidth]{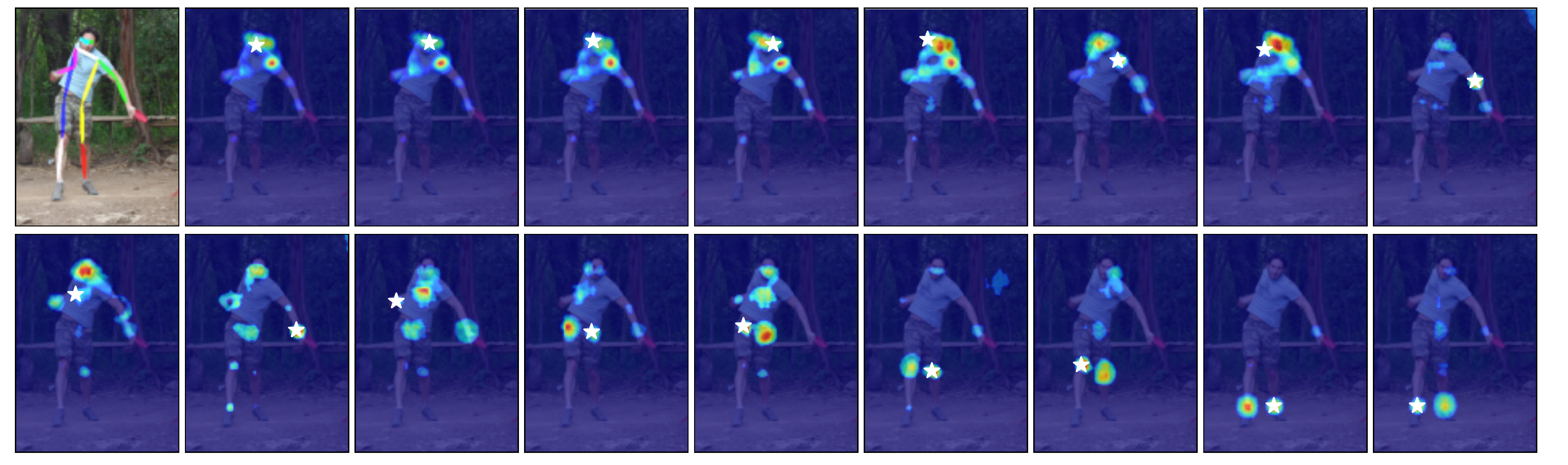}
		\label{tph-a}
	}
	\quad
	
	\hspace{-0.5em}
	\subfigure[\textbf{TP-R-A4:} predicted keypoints and their \emph{dependency areas} for input \textbf{B}.]{
		\includegraphics[width=0.48\linewidth]{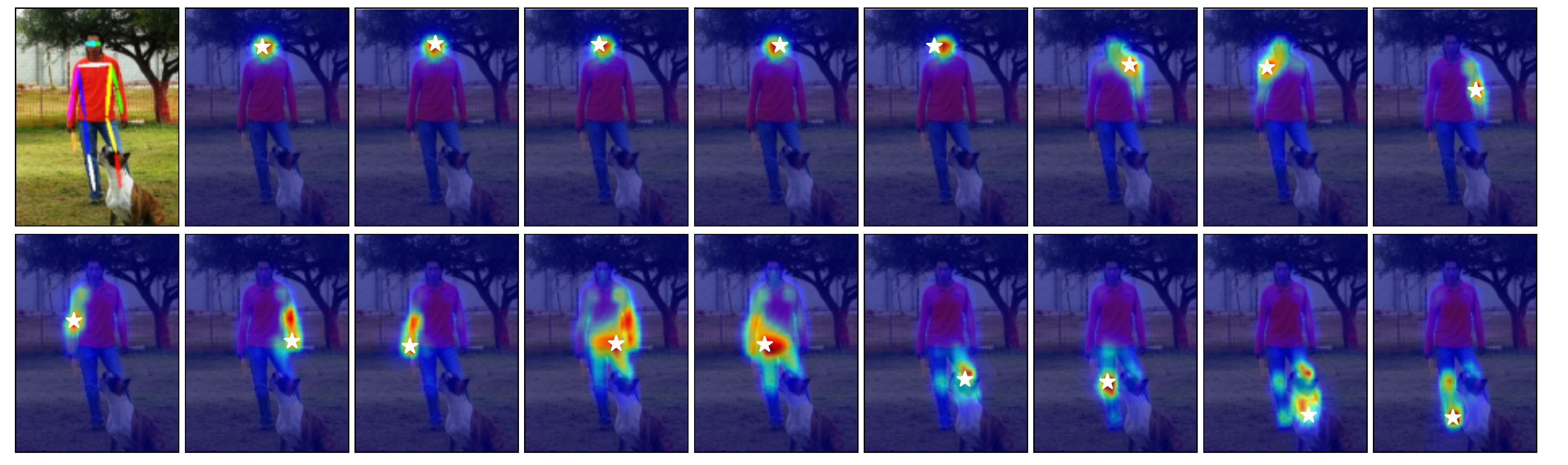}
		\label{tpr-b}
	}\hspace{1.em}\vspace{-0.01in}
	\subfigure[\textbf{TP-H-A4:} predicted keypoints and their \emph{dependency areas} for input \textbf{B}.]{
		\includegraphics[width=0.48\linewidth]{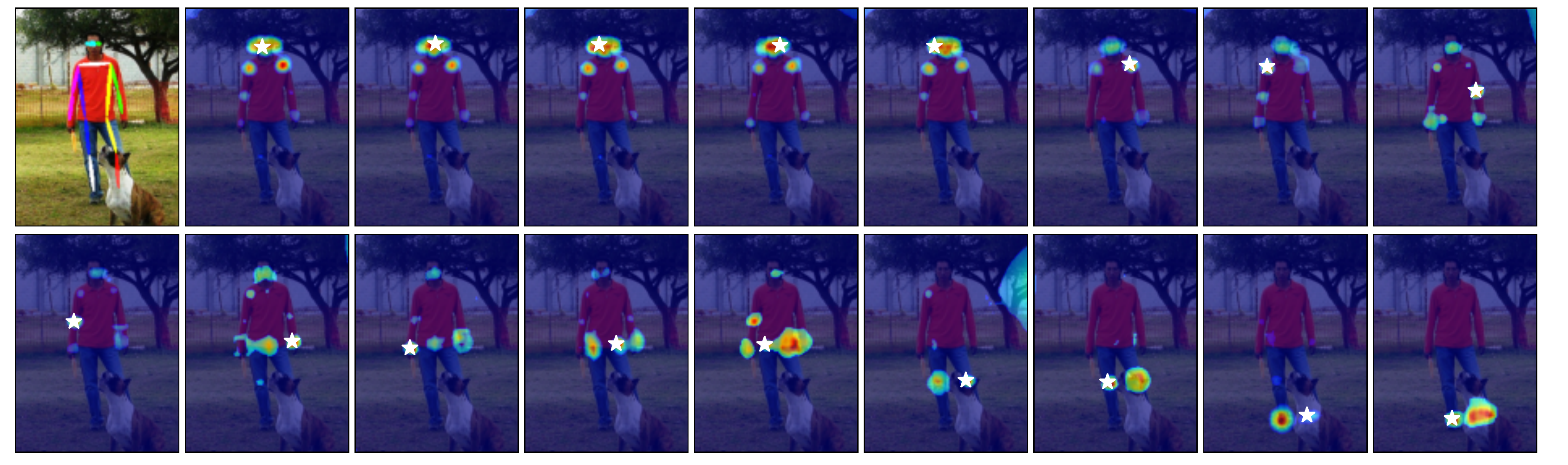}
		\label{tph-b}
	}
	\caption{Predicted locations and the dependency areas for different types of keypoints by different models: TP-R-A4 (left column) and TP-H-A4 (right column). In each sub-figure, the first one is the original input image plotted with predicted skeleton. The other maps visualized by the defined dependency area ($A_{i,:}$) of the attention matrix in the last layer with a threshold value (0.00075). The predicted location of a keypoint is annotated by a WHITE color pentagram ($\star$) in each sub-map. Redder area indicates higher attention scores.}\vspace*{-0.1in}
	\label{last-attention-dependency}
\end{figure*}

\begin{figure*}
	\centering
	\hspace{-1em}
	\subfigure[\textbf{TP-R-A4:} predictions and dependency areas for \textbf{Input C}.]{\includegraphics[width=0.49\linewidth]{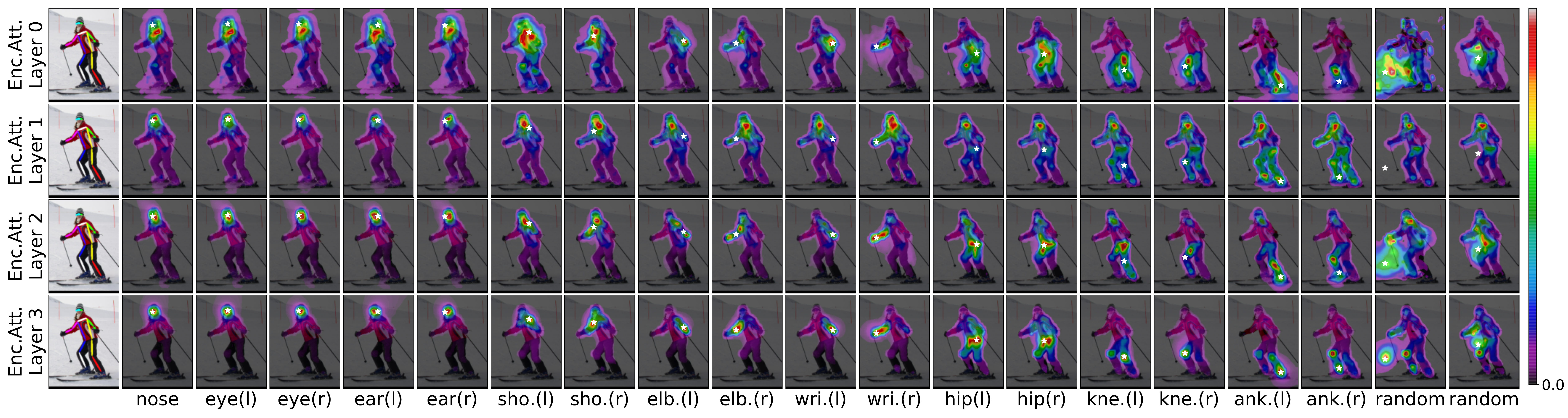}}\hspace{1em}\label{da_tpr} 
	\subfigure[\textbf{TP-H-A4:} predictions and dependency areas for \textbf{Input C}.]{\includegraphics[width=0.49\linewidth]{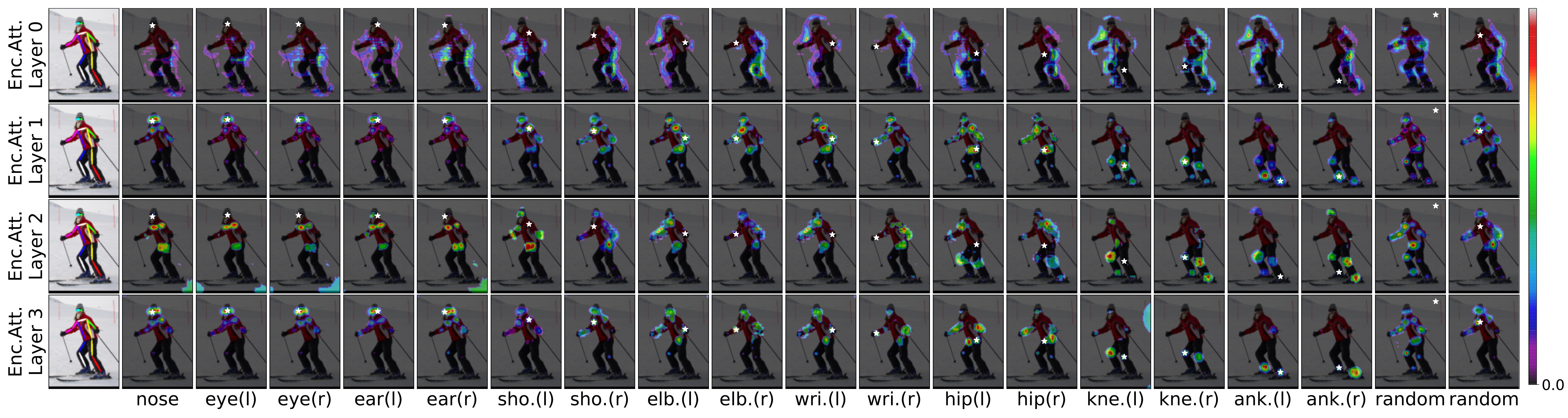}}\label{da_tph}
	
	\caption{\textbf{Dependency areas} for the particular positions in the different attention layers by the same visualization method of Fig.\ref{last-attention-dependency}.}\hspace{1em}\vspace*{-0.1in}
	\label{all-attention-dependency}
\end{figure*}

\subsection{Qualitative Analysis} 

\label{Explainability Analysis}
The hyperparameter configurations for TransPose model might affect the its behavior in an unknown way. In this section, we choose \emph{trained models, types of predicted keypoints, depths of attention layers, and input images} as controlled variables to observe the model behaviors. 

{\bf The dependency preferences are different for models with different CNN extractors.} To make comparisons between ResNet-S and HRNet-S based models, we use the trained models TP-\textbf{R}-A4 and TP-\textbf{H}-A4 performances as exemplars. Illustrated in Fig.~\ref{last-attention-dependency}, we choose two typical inputs A and B as examples and visualize the \emph{dependency areas} defined in Sec.~\ref{paper::definition}. We find that although the predictions from TP-R-A4 and TP-H-A4 are exactly the same locations of keypoints, TP-\textbf{H}-A4 can exploit multiple longer-range joints clues to predict keypoints. In contrast, TP-\textbf{R}-A4 prefers to attend to local image cues around the target joint. This characteristic can be further confirmed by the visualized affected areas in supplementary, in which keypoints have larger and non-local affected areas in TP-\textbf{H}-A4. Although such results are not as commonly expected, they reflect: 1) a pose estimator uses global information from long-range joints to localize a particular joint; 2) HRNet-S is better than ResNet-S at capturing long-range dependency relationships information (probably due to its multi-scale fusion scheme).

{\bf Dependencies and influences vary for different types of keypoints.} For keypoints in the head, localizing them mainly relies on visual clues from head, but TP-\textbf{H}-A4 also associates them with shoulders and the joints of arms. Notably, the dependencies of predicting wrists, elbows, knees or ankles have obvious differences for two models, in which TP-\textbf{R}-A4 depends on the local clues at the same side while TP-\textbf{H}-A4 exploits more clues from the joints on the symmetrical side. As shown in Fig.~\ref{tph-a}, Fig.~\ref{tph-b}, and Fig.~\ref{all-attention-dependency}, we can further observe that a pose estimator might gather strong clues from more parts to predict the target keypoint.  This can explain why the model still can predict the location of an occluded keypoint accurately, and the occluded keypoint with ambiguity location will have less impact on the other predictions or larger uncertain area to rely on (e.g. the occluded left ankle -- last map of Fig.~\ref{tpr-b} or Fig.~\ref{tph-b}).

{\bf Attentions gradually focus on more fine-grained dependencies with the depth increasing.}  Observing all of attention layers (the 1,2,3-th rows of Fig.~\ref{all-attention-dependency}), we surprisingly find that \emph{even without the intermediate GT locations supervision}, TP-H-A4 can still attend to the accurate locations of joints yet with more global cues in the early attention layers. For both models, with the depth increasing, the predictions gradually depend on more fine-grained image clues around local parts or keypoints positions (Fig.~\ref{all-attention-dependency}).

{\bf Image-specific dependencies and statistical commonalities for a single model.} Different from the static relationships encoded in the weights of CNN after training, the attention maps are dynamic to inputs. As shown in Fig.~\ref{tpr-a} and Fig.~\ref{tpr-b}, we can observe that despite the statistical commonalities on the dependency relationships for the predicted keypoints (similar behaviors for most common images), the fine-grained dependencies would slightly change according to the image context. With the existence of occlusion or invisibility in a given image such as input B (Fig.~\ref{tpr-b}), the model can still localize the position of the partially obscured keypoint by looking for more significant image clues and reduces reliance on the invisible keypoint to predict the other ones. It is likely that future works can exploit such attention patterns for parts-to-whole association and aggregating relevant features for 3D pose estimation or action recognition. 

\section{Conclusion}
We explored a model -- TransPose -- by introducing Transformer for human pose estimation. The attention layers enable the model to capture global spatial dependencies efficiently and explicitly. And we show that such a heatmap-based localization achieved by Transformer makes our model share the idea with Activation Maximization. 
With lightweight architectures, TransPose matches state-of-the-art CNN-based counterparts on COCO and gains significant improvements on MPII when fine-tuned with small training costs. Furthermore, we validate the importance of position embedding. Our qualitative analysis reveals the model behaviors that are variable for layer depths, keypoints types, trained models and input images, which also gives us insights into how models handle special cases such as occlusion.

\textbf{Acknowledgment}. 
This work was supported by the National Natural Science Foundation of China (61773117 and 62006041).



{
	\bibliographystyle{ieee_fullname}
	\bibliography{egbib}
}

\newpage
\appendix
\section{2D Sine Position Embedding}
\label{appendix::2dpe}
Without the position information embedded in the input sequence, the Transformer Encoder is a permutation-equivariant architecture:
\begin{equation}
\operatorname{Encoder}\left(\rho \left(\mathbf{X} \right)\right) = \rho \left(\operatorname{Encoder}\left(\mathbf{X} \right)\right),
\end{equation}
where $\rho$ is any permutation for the pixel locations or the order of sequence. To
make the order of sequence or the spatial structure of the image pixels
matter, we follow the sine positional encodings but further hypothesize that the position information is independent at $x$ (horizontal) and $y$ (vertical) direction of an image, like the ways of~\cite{parmar2018image, carion2020detr}. Concretely, we keep the original 2D-structure respectively with $d/2$ channels for $x,y$-direction:
\begin{equation}
\begin{aligned}
P E_{(2i, p_y, :)} &=\sin \left( 2\pi* p_y /(H*10000^{2i / \frac{d}{2}}) \right), \\
P E_{(2 i+1, p_y,:)} &=\cos \left( 2\pi* p_y /(H*10000^{2i / \frac{d}{2}} )\right),  \\
P E_{( 2i,: ,p_x)} &=\sin \left( 2\pi* p_x /(W*10000^{2i / \frac{d}{2}} )\right), \\
P E_{( 2 i+1,:,p_x)} &=\cos \left( 2\pi* p_x /(W*10000^{2i / \frac{d}{2}} )\right),
\end{aligned}
\end{equation}
where $i=0,1,...,d/2-1$, $p_x$ or $p_y$ is the position index along $x$ or $y$-direction. Then they are stacked and flattened into a shape $\mathbb{R}^{ L\times d}$. The position embedding is injected into the input sequences before self-attention computation. We use 2D sine position embedding by default for all models. 

\section{What position information has been learned in the TransPose model with learnable position embedding?}

\begin{figure}[h]
	\begin{center}
		\includegraphics[height=0.75\linewidth]{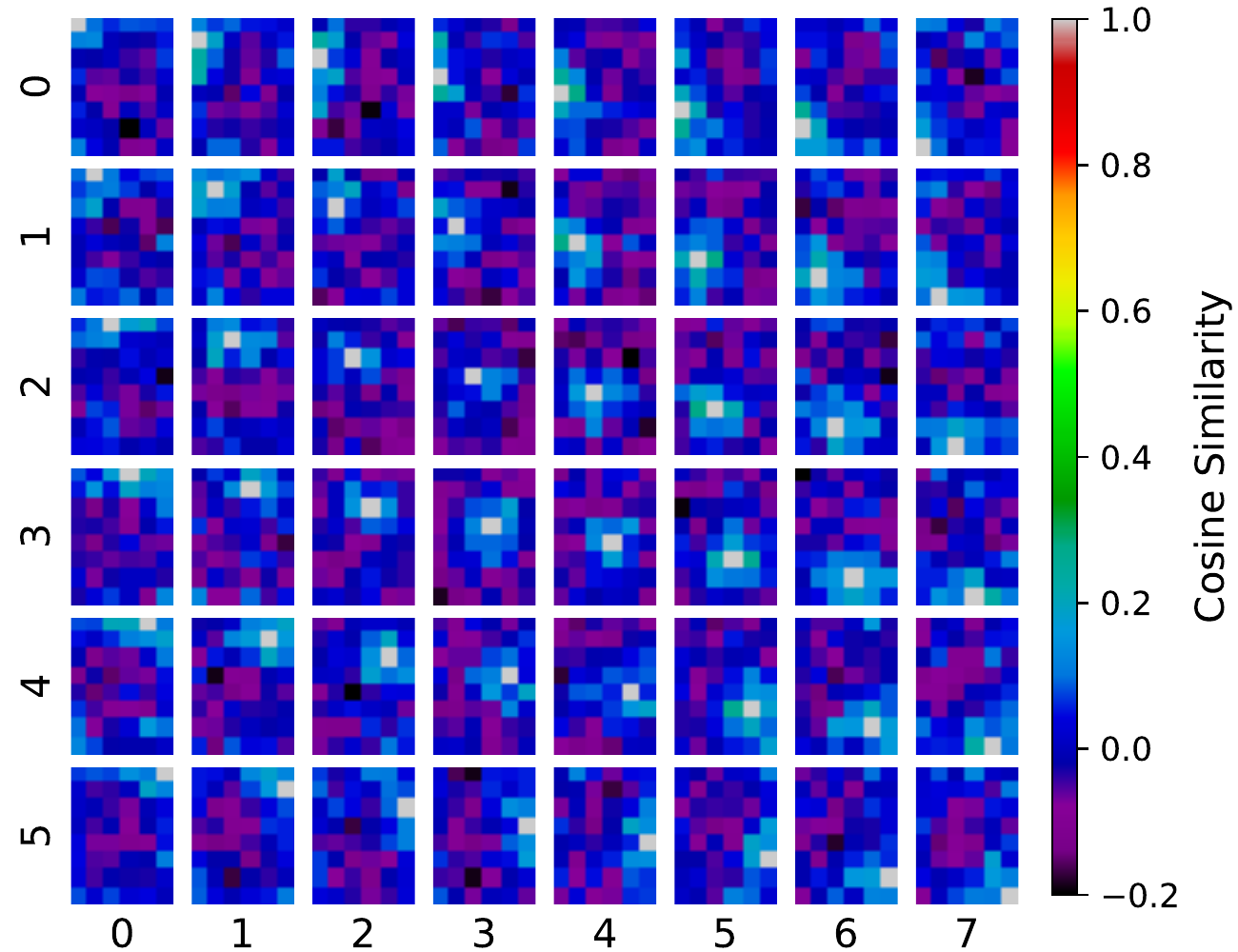}
	\end{center}
	\caption{The cosine similarities between the learned position embedding vectors, which have been reshaped into 2D grid and interpolated with 0.25 scale factor for a better illustration (the original shape is $\left(24, 32\right) $). Each map in $x$-row and $y$-col of the figure represents the cosine similarities between the embedding vector in position $(x,y)$ and the embedding vectors at other locations.}
	\label{appendix::lpe}
\end{figure}

We show what position information has been learned in the TransPose (TransPose-R) with learnable position embedding. It has been discussed in the paper. As shown in Fig.~\ref{appendix::lpe}, we visualize the similarities by calculating the cosine similarity between vectors at any pair of locations of the learnable position embedding and reshaping it into a 2D grid-like map. We find that the embedding in each location of learnable position embedding has a unique vector value in the $d$-dim vector space, but it has relatively higher cosine similarity values with the neighbour locations in 2D-grid and lower values with those far away from it. The results indicate the coarse 2D position information has been implicitly learned in the learnable position embedding. We suppose that the learning sources of the position information might be the 2D-structure groundtruth heatmaps and the similar features existing in the 1D-structure sequences. The model learns to build associations between position embedding and input sequences, as a result it can predict the target heatmaps with 2D Gaussian peaking at groundtruth keypoints locations.

In the paper, we find that position embedding helps to generalize better on unseen input resolutions, particularly 2D sine position embedding. We conjecture that 1) the models with a fixed receptive field may be hard to adapt the changes in scales; 2) building associations with \emph{position information} encoded in Sine position embedding~\cite{vaswani2017attention} may help model generalize better on different sizes.

\section{Transformer Encoder Layer}
\label{appendix::encoder}
The Transformer Encoder layer~\cite{vaswani2017attention} we used can be formulated as:
\begin{equation}
\begin{aligned}
\mathbf{Z}=&\operatorname{LayerNorm}\left( \operatorname{MultiheadSelfAttention}\left( \mathbf{X}\right) + \mathbf{X} \right),  \\
\mathbf{X^*}=&\operatorname{LayerNorm}\left( \operatorname{FFN}\left( \mathbf{Z}\right) + \mathbf{Z} \right),
\end{aligned}
\end{equation}
where $\mathbf{X}$ is the original input sequence that has not yet been added with position embedding. The position embedding will be added to $\mathbf{X}$ for computing querys and keys excluding values. $\mathbf{X^*}$ is the output sequence of the current Transformer Encoder layer, as the input sequence of next encoder layer.	 The formulations of Multihead Self-Attention and FFN are defined in~\cite{vaswani2017attention}.

\section {Gradient Analysis}
\label{appendix::grad} 
From the view of an activation at some location $i$ of the predicted heatmaps, the network weights associating all input tokens across the whole image/sequence with this activation can be seen as a discriminator that judges the presence or absence of a certain keypoint at this location. As revealed by~\cite{samek2019explainable,simonyan2013deep, bach2015pixel, selvaraju2017grad}, the gradient information can indicate the importance (sensitivity) of the input features to a specific output of a non-linear model. That assumption is based on that tiny change in the input (pixel/feature/token) with the most important feature value causes a large change in what the output of the model would be. 

Suppose we have a trained model and a specific image, $\boldsymbol{h}_i\in \mathbb{R}^{K}$ is the scores for all $K$ types of keypoints at location $i$ of the predicted heatmaps; $\boldsymbol{z}_i\in \mathbb{R}^d$ is the intermediate feature outputted by the last self-attention layer before being fed into FFN. There is only a ReLU excluding the linear and convolutions\footnote{a $1\times1$ convolution is also a position-wise linear layer; the $4\times4$ deconvolution used in TP-R acts as the upsampling operation.} (head) layers after the last attention layer. ReLU (rectified linear unit) activation function in FFN can be empirically regarded as a negative contribution filter, which only retains positive contributions and maintains the linearity. Next, we choose numerator layout for computing the derivative of a vector with respect to a vector.  We thus assume the mapping from $\boldsymbol{z}_i$ to $\boldsymbol{h}_i$ can be approximated as a linear function $f$ with learned weights $\mathbf{W}_f\in \mathbb{R}^{K \times d}$ and bias $\mathbf{b}\in\mathbb{R}^K$ by computing the first-order Taylor expansion at a given local point $\boldsymbol{z}_i^0$, \emph{i.e.}, $\boldsymbol{h}_i\approx\mathbf{W}_f\boldsymbol{z}_i+\mathbf{b}$, $\mathbf{W}_f=\left.\frac{\partial \boldsymbol{h}_i}{\partial \boldsymbol{z}_i}\right|_{\boldsymbol{z}_i^0}$.   Then we compute the partial derivative of $\boldsymbol{h}_i$ at location $i$ of the output heatmaps w.r.t the token $\boldsymbol{x}_j$ at location $j$ of the input sequence of the last attention layer:
\begin{equation}
\begin{aligned}
\frac{\partial \boldsymbol{h}_i}{\partial \boldsymbol{x}_j}&=\frac{\partial \boldsymbol{h}_i}{\partial \boldsymbol{z}_i}\frac{\partial \boldsymbol{z}_i}{\partial \boldsymbol{x}_j}\\
&=\frac{\partial f(\boldsymbol{z}_i)}{\partial \boldsymbol{z}_i}(\boldsymbol{1}+\frac{\partial \mathbf{w}_i\mathbf{V}}{\partial \boldsymbol{x}_j})\\
&\approx \mathbf{W}_f(\boldsymbol{1}+\frac{\partial \mathbf{w}_{i,0}\boldsymbol{v}_0+...+\mathbf{w}_{i,j}\boldsymbol{v}_j+...+\mathbf{w}_{i,L-1}\boldsymbol{v}_{L-1}}{\partial \boldsymbol{x}_j})\\
&=\mathbf{W}_f(\boldsymbol{1}+\frac{\partial \mathbf{w}_{i,j}\boldsymbol{v}_j}{\partial \boldsymbol{x}_j})\\
&=\mathbf{W}_f(\boldsymbol{1}+\frac{\partial \mathbf{A}_{i,j}\mathbf{W}_v^\top\boldsymbol{x}_j}{\partial \boldsymbol{x}_j})\\
\end{aligned}
\end{equation}
where $\boldsymbol{v}_j\in\mathbb{R}^d$ is the value vector transformed by: $\boldsymbol{v}_j=\mathbf{W}_v^\top\boldsymbol{x}_j$. $\mathbf{A}_{i,j}$ is a scalar value that is computed by the dot-product between $\boldsymbol{q}_i$ and $\boldsymbol{k}_j$. We assume $G :=\frac{\partial \boldsymbol{h}_i}{\partial \boldsymbol{x}_j}$ as a function w.r.t. a given attention score $\mathbf{A}_{i,j}$. Under this assumption $\mathbf{A}_{i,j}$ is deemed as an observed variable that has blocked its parent nodes. Then we define:
\begin{equation}
\begin{aligned}
G\left(\mathbf{A}_{i,j} \right) &=\mathbf{W}_f(\boldsymbol{1}+\frac{\partial \mathbf{A}_{i,j}\mathbf{W}_v^\top\boldsymbol{x}_j}{\partial \boldsymbol{x}_j})\\
\\
&=\mathbf{W}_f\left( \boldsymbol{1}+\mathbf{A}_{i,j}\mathbf{W}_v^\top\right) \\
&=\mathbf{A}_{i,j}\mathbf{W}_f\mathbf{W}_v^\top+\mathbf{W}_f\\
&=\underbrace{\mathbf{A}_{i,j}}_{\textbf{Image-Specific: dynamic weights}}\cdot\underbrace{\mathbf{W}_f
	\cdot\mathbf{W}_v^\top+\mathbf{W}_f}_{\textbf{Learned: static weights}}\\
&=\mathbf{A}_{i,j}\cdot\mathbf{K} + \mathbf{B}
\end{aligned}
\end{equation}
where $\mathbf{K},\mathbf{B}\in\mathbb{R}^{K\times d}$ are static weights shared across all positions.  We can see that the function $G$ is approximately linear with $\mathbf{A}_{i,j}$, \emph{i.e.}, the degrees of contribution to the prediction $\boldsymbol{h}_i$ directly depend on its attention scores at those locations.

The last attention layer in Transformer Encoder, whose attention scores are seen as the image-specific weights, aggregate contributions from all locations according to attention scores and finally form the maximum activations in the output heatmaps. Though the layers in FFN and head cannot be ignored\footnote{1. Assuming that the used convolutions extract feature in a limited patch, the global interactions mostly occur at the attention layers. 2. The layer normalization does not affect the interactions between locations.}, they are position-wise operators, which almost linearly transform the attention scores from all the positions with the same transformation. In addition, $\mathbf{Q}=\left( \mathbf{X}+\mathbf{P}\right) \mathbf{W}_q,\mathbf{K}=\left( \mathbf{X}+\mathbf{P}\right)\mathbf{W}_k,\mathbf{V}=\mathbf{X}\mathbf{W}_v$ where $\mathbf{P}$ is the position embedding. Because $\mathbf{A}_{i,j}\propto\mathbf{Q}_i\mathbf{K}_j^\top$, the position embedding values also affect the attention scores to some extent.

\begin{figure*}[h]
	\centering
	\subfigure[\textbf{TP-R-A4:} predictions and dependency areas for \textbf{input 1}.]{
		\includegraphics[width=0.475\linewidth]{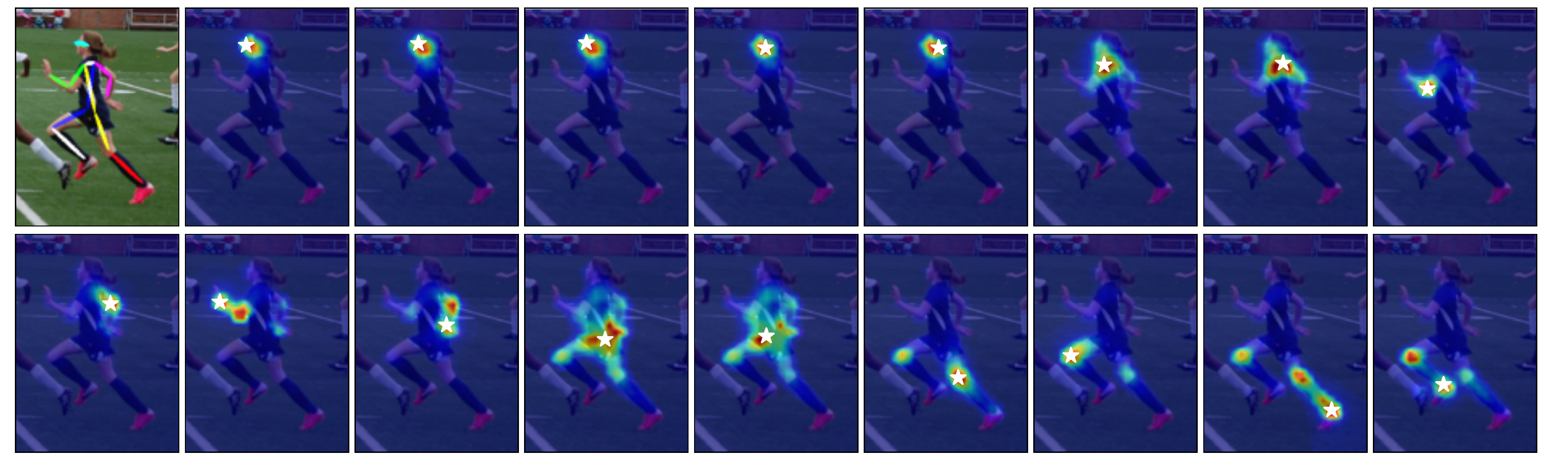}
		
	}
	\quad
	\subfigure[\textbf{TP-H-A4:} predictions and dependency areas for \textbf{input 1}.]{
		\includegraphics[width=0.475\linewidth]{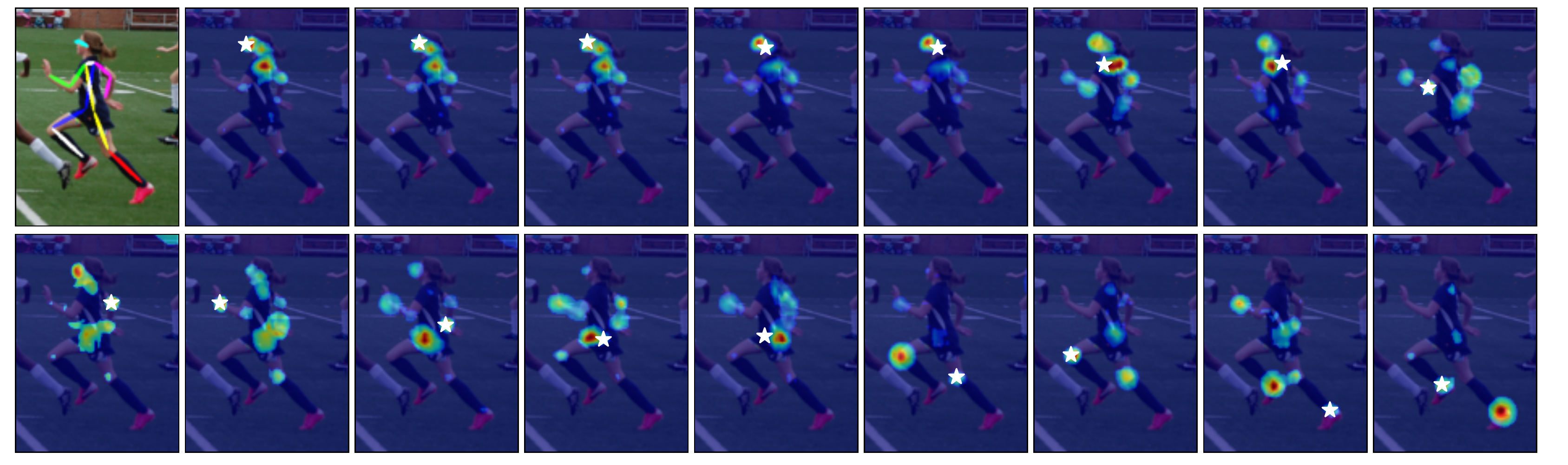}
	}
	
	\subfigure[\textbf{TP-R-A4:} predictions and dependency areas for \textbf{input 2}.]{
		\includegraphics[width=0.475\linewidth]{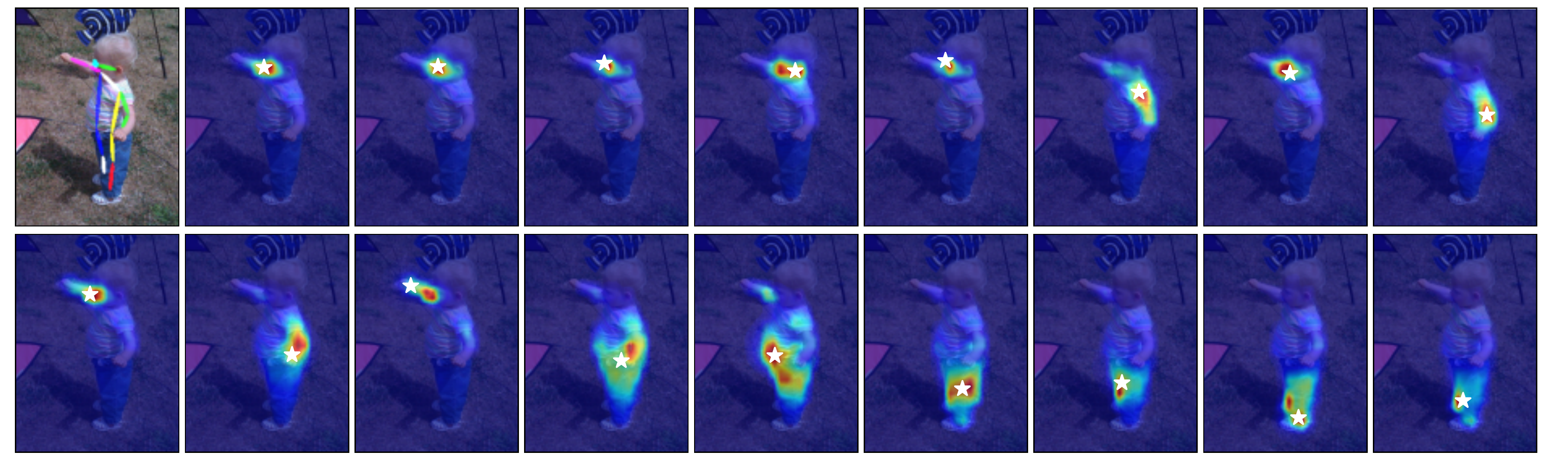}
		
	}
	\quad
	\subfigure[\textbf{TP-H-A4:} predictions and dependency areas for \textbf{input 2}.]{
		\includegraphics[width=0.475\linewidth]{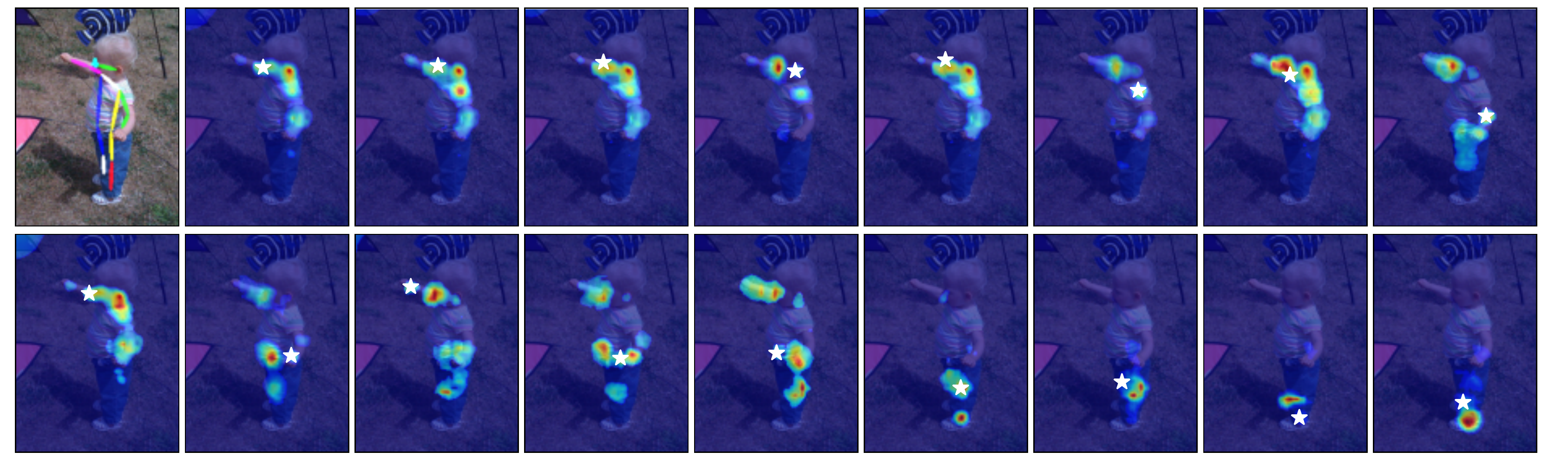}
	}
	
	\subfigure[\textbf{TP-R-A4:} predictions and dependency areas for \textbf{input 3}.]{
		\includegraphics[width=0.475\linewidth]{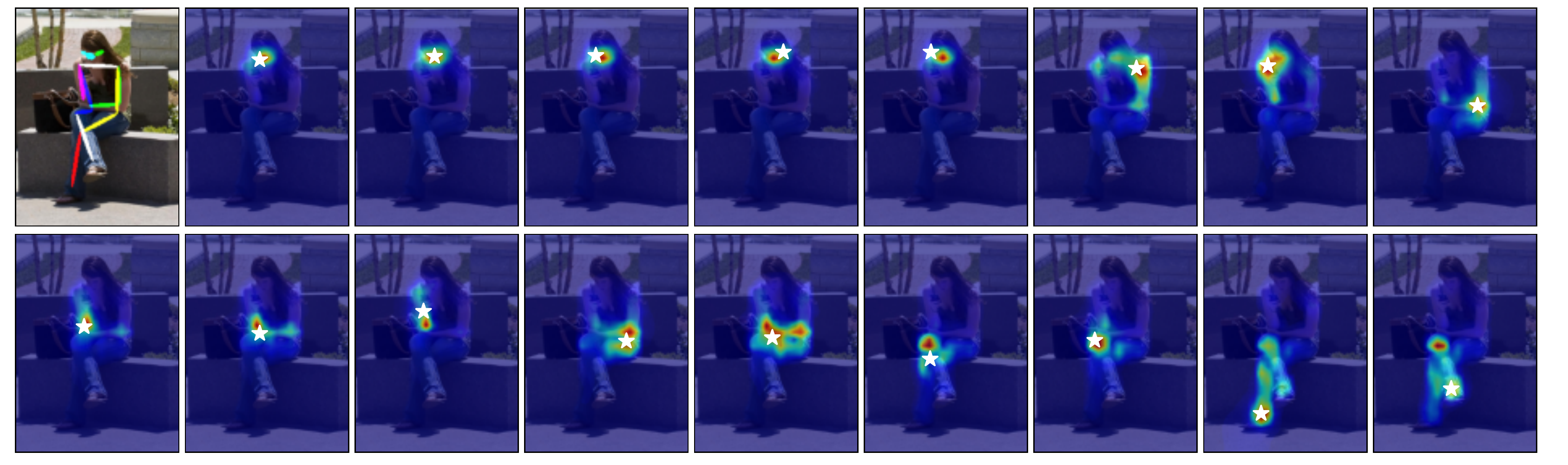}
		
	}
	\quad
	\subfigure[\textbf{TP-H-A4:} predictions and dependency areas for \textbf{input 3}.]{
		\includegraphics[width=0.475\linewidth]{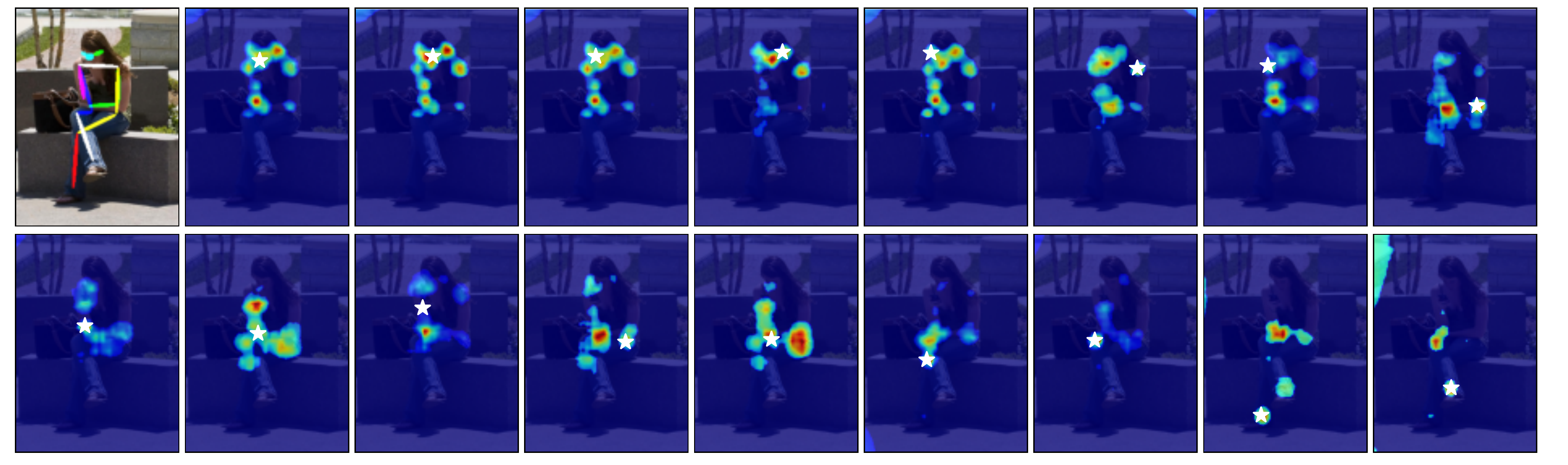}

	}
	\quad
	\subfigure[\textbf{TP-R-A4:} predictions and dependency areas for \textbf{input 4}.]{
		\includegraphics[width=0.475\linewidth]{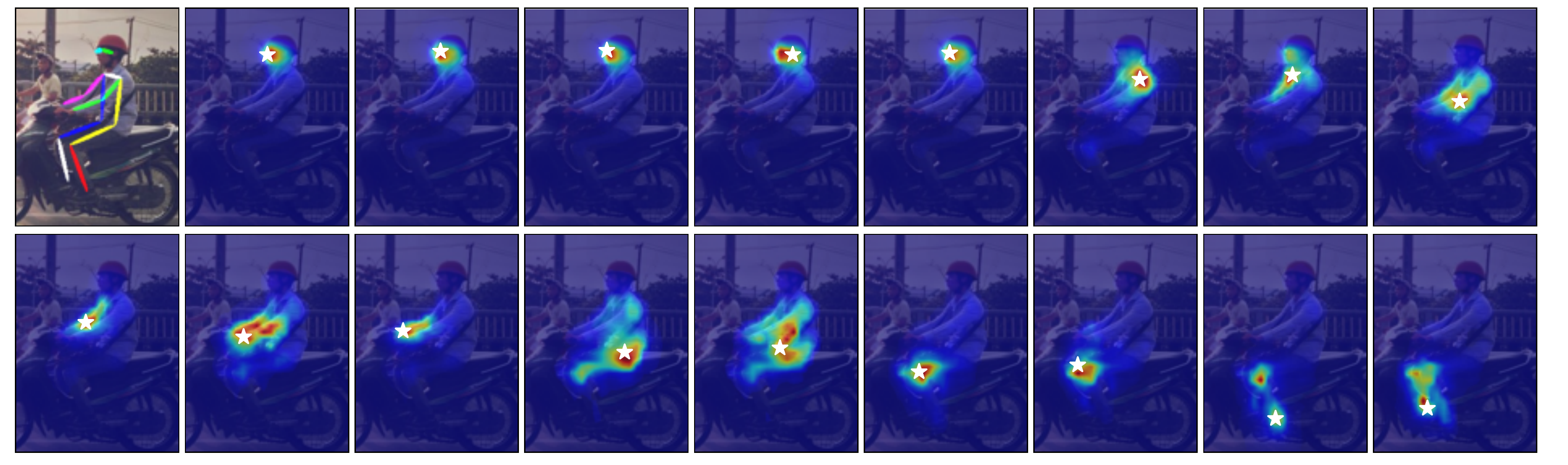}
		
	}
	\quad
	\subfigure[\textbf{TP-H-A4:} predictions and dependency areas for \textbf{input 4}.]{
		\includegraphics[width=0.475\linewidth]{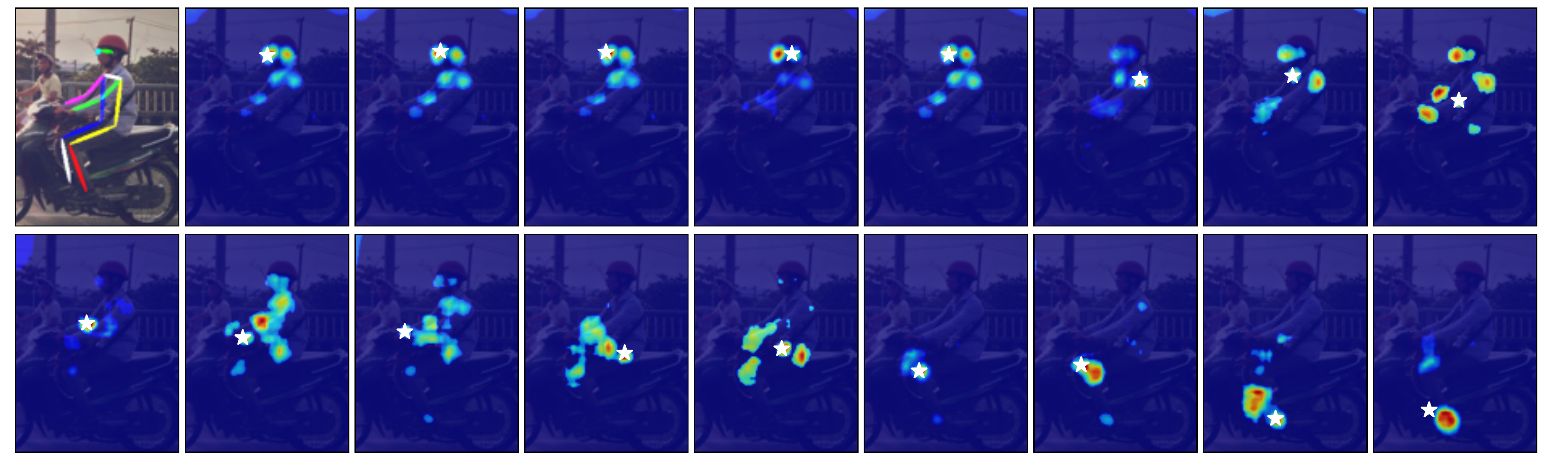}
		\label{tph-x}
	}

	\subfigure[\textbf{TP-R-A4:} predictions and dependency areas for \textbf{input 5}.]{
		\includegraphics[width=0.475\linewidth]{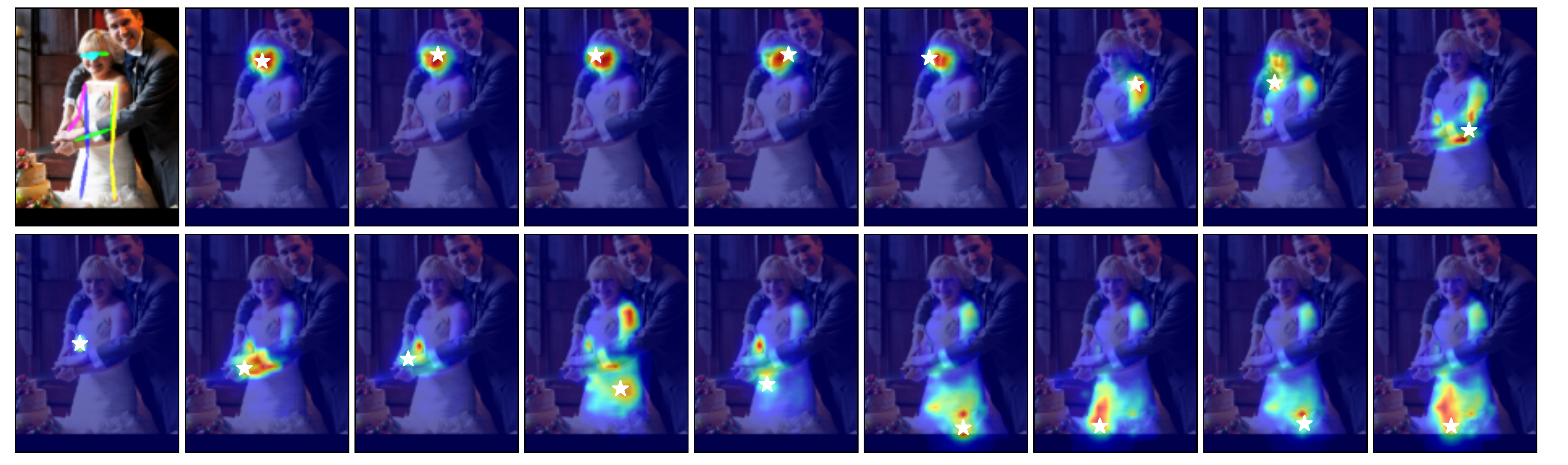}
		
	}
	\quad
	\subfigure[\textbf{TP-H-A4:} predictions and dependency areas for \textbf{input 5}.]{
		\includegraphics[width=0.475\linewidth]{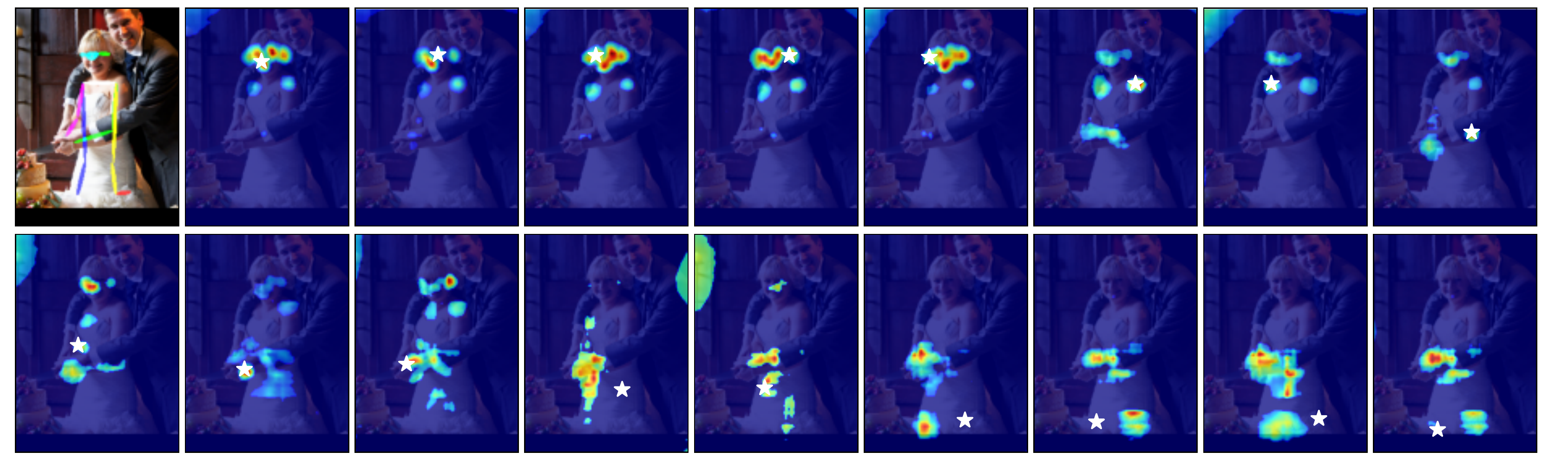}
	}
	
	\subfigure[\textbf{TP-R-A4:} predictions and dependency areas for \textbf{input 5}.]{
		\includegraphics[width=0.475\linewidth]{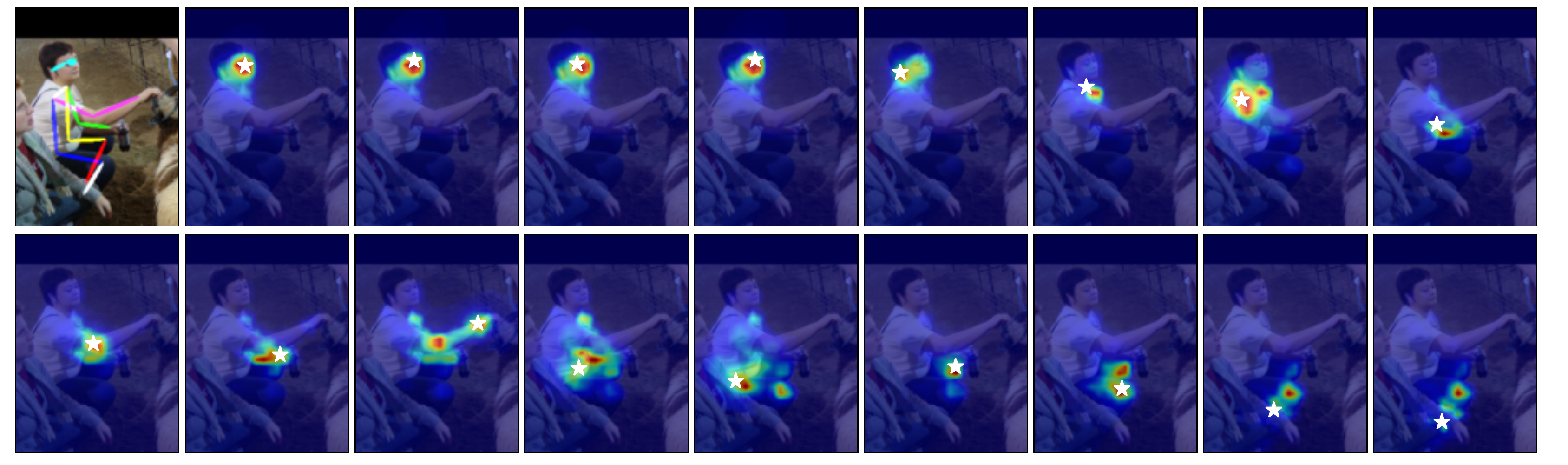}
		
	}
	\quad
	\subfigure[\textbf{TP-H-A4:} predictions and dependency areas for \textbf{input 5}.]{
		\includegraphics[width=0.475\linewidth]{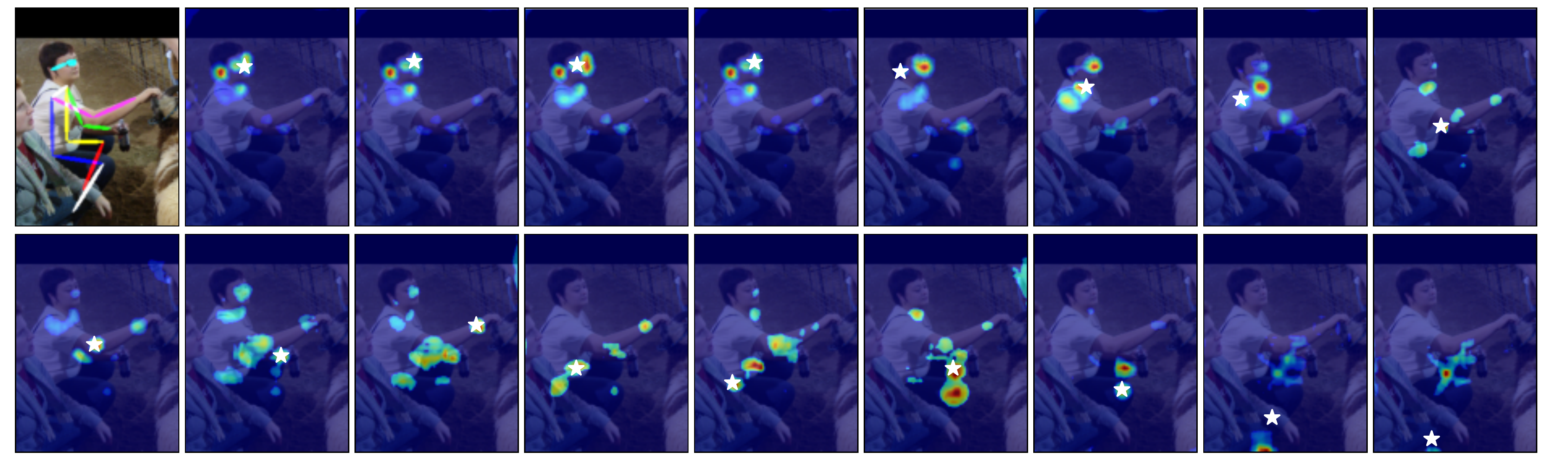}
	}

	\caption{Predicted locations and the dependency areas for different types of keypoints in different models: TP-R-A4 (left column) and TP-H-A4 (right column). In each sub-figure, the first one is the original input image plotted with predicted skeleton. The other maps visualized by the defined dependency area ($A_{i,:}$) of the attention matrix in the last layer with a threshold value (0.00075). The predicted location of a keypoint is annotated by a WHITE color pentagram ($\star$) in each sub-map. Redder area indicates higher attention scores.}
	\label{appendix::final_attention}
\end{figure*}

\begin{figure*}
	\centering
	\subfigure[\textbf{TP-R-A4:} predictions and dependency areas of each keypoint in different attention layers.]{
		\includegraphics[width=0.45\linewidth]{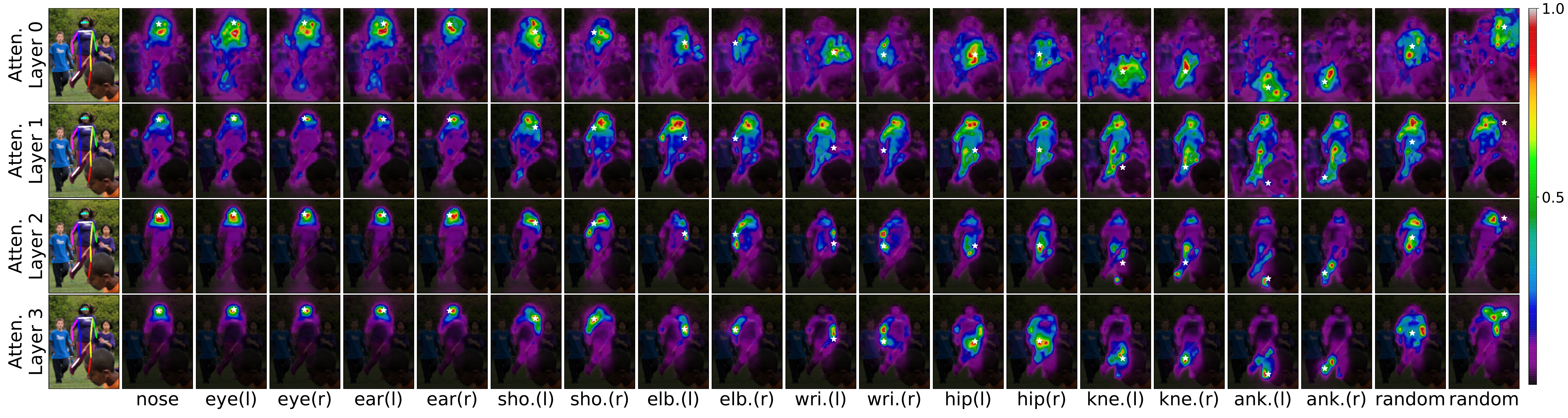}
	}
	\quad
	\subfigure[\textbf{TP-H-A4:} predictions and dependency areas of each keypoint in different attention layers.]{
		\includegraphics[width=0.45\linewidth]{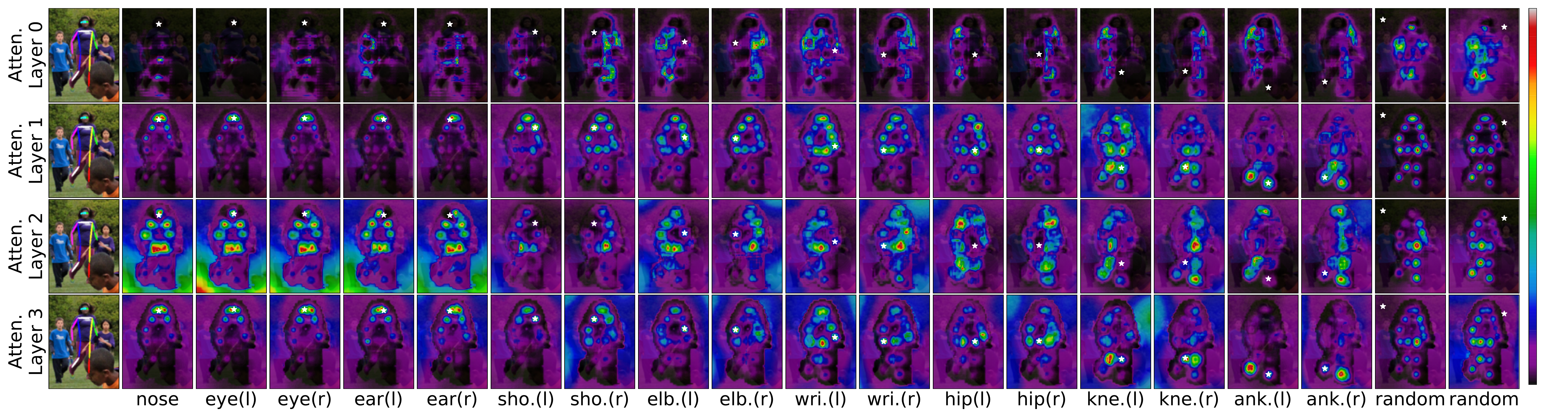}
	}
	\subfigure[\textbf{TP-R-A4:} predictions and dependency areas of each keypoint in different attention layers.]{
		\includegraphics[width=0.45\linewidth]{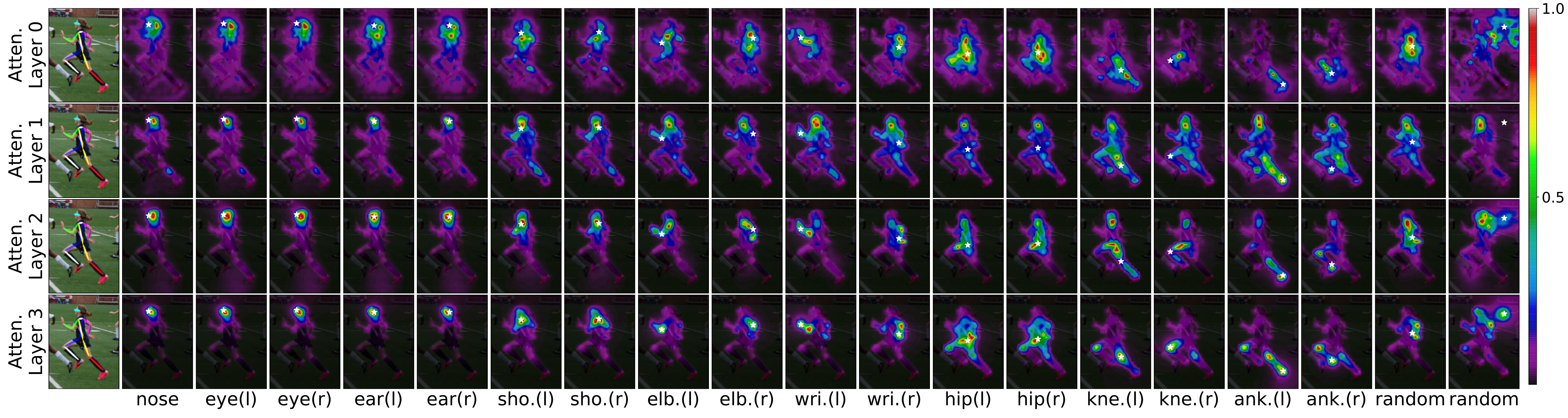}
	}
	\quad
	\subfigure[\textbf{TP-H-A4:} predictions and dependency areas of each keypoint in different attention layers.]{
		\includegraphics[width=0.45\linewidth]{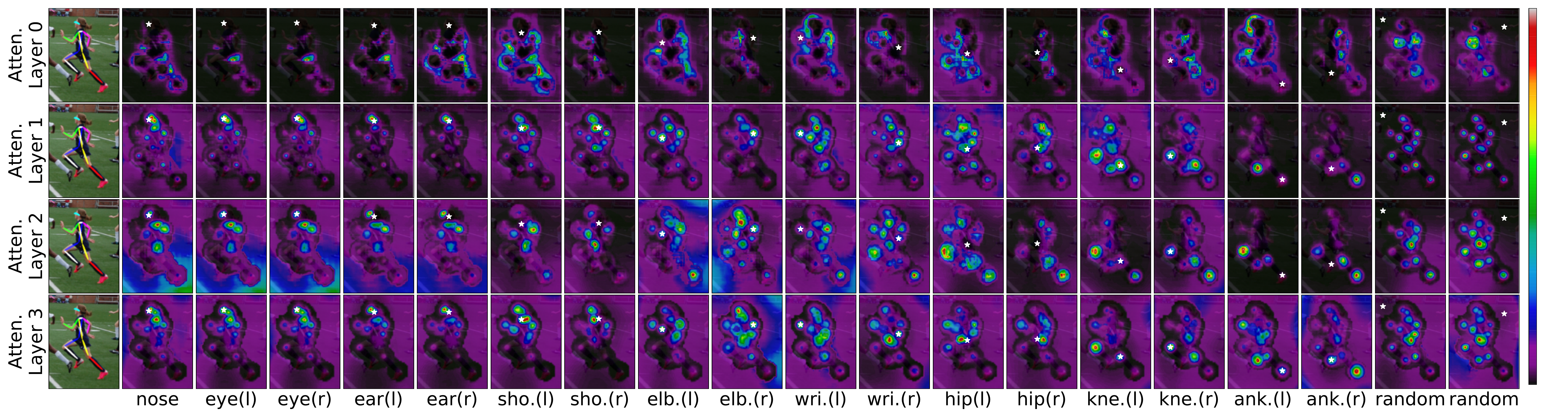}
	}
	\subfigure[\textbf{TP-R-A4:} predictions and affect areas of each keypoint in different attention layers.]{
		\includegraphics[width=0.45\linewidth]{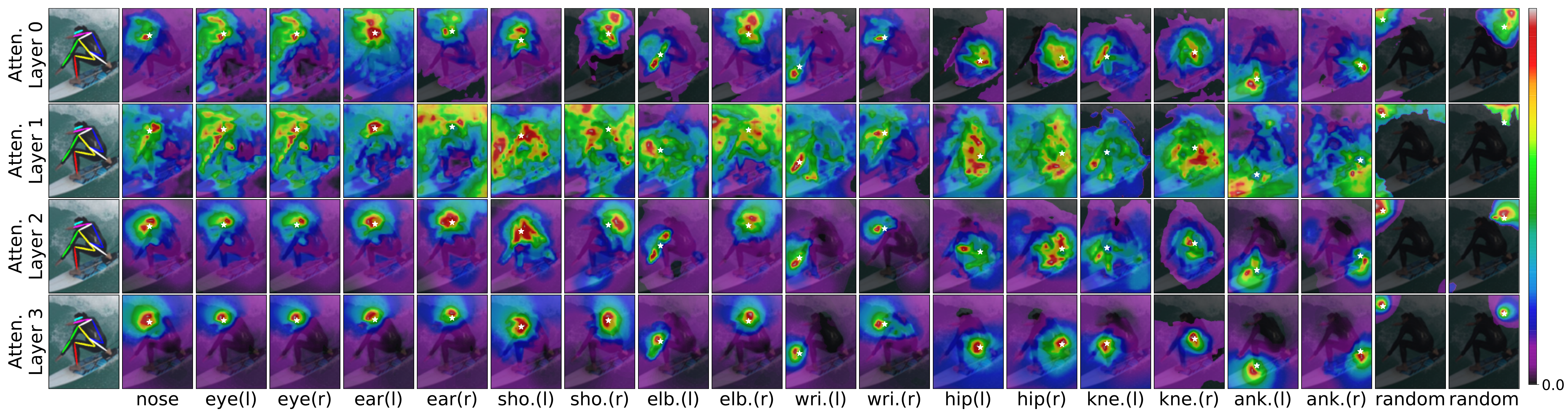}
	}
	\quad
	\subfigure[\textbf{TP-H-A4:} predictions and affect areas of each keypoint in different attention layers.]{
		\includegraphics[width=0.45\linewidth]{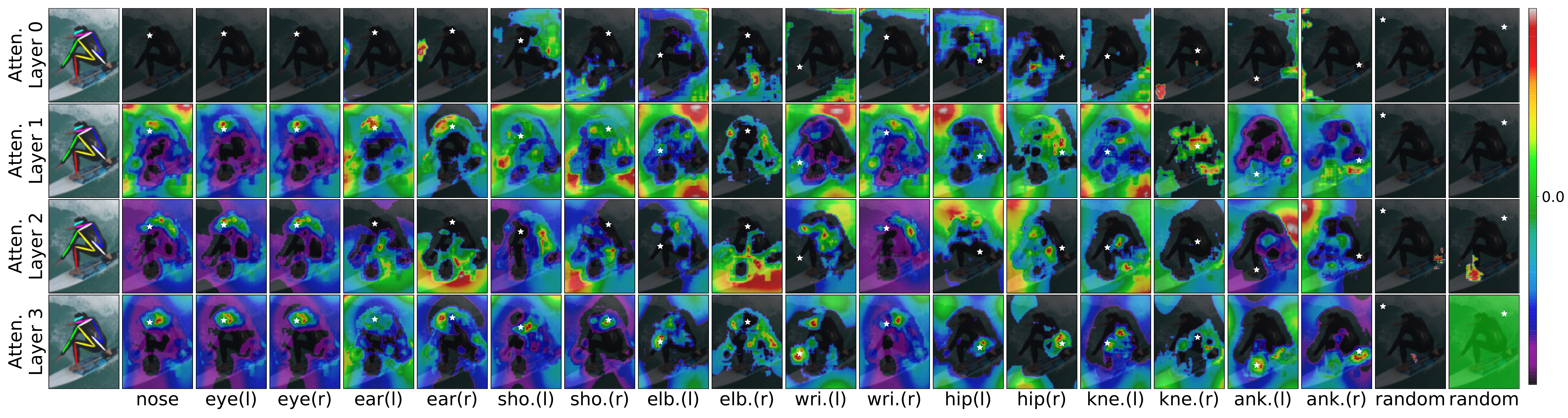}
	}
	\caption{\textbf{Dependency areas} (the first two rows) and \textbf{Affected areas} (the last row) in different attention layers for different input images.}
	\label{appendix::dependency-and-affect}
\end{figure*}

\begin{table}
	\renewcommand{\arraystretch}{0.9}
	\centering
	
	\begin{tabular}{c|c}
		\toprule[0.1em]
		
		Backbone &ResNet-S\\
		\midrule
		\multirow{2}{*}{Stem}&Conv-k7-s2-c64, BN, ReLU\\
		&Pooling-k3-s2\\
		\hline
		\multirow{4}{*}{Blocks}& 3$\times$Bottleneck-c64\\
		&  Bottleneck-s2-c128 \\
		&  3$\times$Bottleneck-c128\\
		& Conv-k1-s1-c256 \\
		
		\bottomrule[0.1em]
	\end{tabular}
	\caption{The detailed configurations for ResNet-S. Conv-k7-s2-c64 means a convolutional layer with 7$\times$7 kernel size, 2 stride, and 64 output channels, followed by a BN and ReLU; the same below. The Bottleneck-c64 includes Conv-k1-s1-c64-BN-ReLU, Conv-k3-s1-c64-BN-ReLU, and Conv-k1-s1-c256-BN. Bottleneck-c128 includes Conv-k1-s1-c128-BN-ReLU, Conv-k3-s1-c128-BN-ReLU, and Conv-k1-s1-c512-BN. See details in~\cite{he2016deep}.}
	\label{resnet}
\end{table}

\begin{table}
	\renewcommand{\arraystretch}{0.9}
	\centering
	\begin{tabular}{c|c}
		\toprule[0.1em]
		
		Backbone &HRNet-S-W32(48)\\
		\midrule
		\multirow{3}{*}{Stem}&Conv-k3-s2-c64, BN, ReLU\\
		&Conv-k3-s2-c64, BN, ReLU\\
		&4$\times$Bottleneck-c64\\
		\hline
		\multirow{3}{*}{Blocks}& transition1$\sim$stage2\\
		&  transition2$\sim$stage3 \\
		& Conv-k1-s1-c64(92) \\
		
		\bottomrule[0.1em]
	\end{tabular}
	\caption{The detailed configurations for HRNet-S-W32(48). More detailed information about the transition layer and stage blocks are described in the HRNet paper~\cite{sun2019hrnet}.}
	\label{hrnet}
\end{table}

\label{cnn_detail}
\section{Architecture Details}
We report the architecture details of ResNet-S and HRNet-S-W32(48) in Tab.~\ref{resnet} and Tab.~\ref{hrnet}. The ResNet-S* only differs from ResNet-S in that ResNet-S* has 10 Bottleneck-c128 blocks. More details about HRNet-W32 and HRNet-W48 are described in~\cite{sun2019hrnet}.

\section{More Attention Maps Visualizations}
In this section, we show more visualization results of the attention maps from TP-R-A4 (TransPose-R-A4) and TP-H-A4 (TransPose-H-A4) models. The attention maps of the last attention layers of two models are shown in Fig.~\ref{appendix::final_attention}. The attention maps in different attention layers of two models are shown in Fig.~\ref{appendix::dependency-and-affect}.

\end{document}